\documentclass[journal]{IEEEtran}
\usepackage{amsmath,amsfonts}
\usepackage{algorithmic}
\usepackage{placeins}
\usepackage{booktabs}
\usepackage{algorithm}
\usepackage{array}
\usepackage{pifont}
\usepackage{xcolor}
\usepackage{comment}
\usepackage{amssymb}
\usepackage[caption=false,font=normalsize,labelfont=normalsize,textfont=normalsize]{subfig}
\usepackage{textcomp}
\usepackage{stfloats}
\usepackage{url}
\usepackage{verbatim}
\usepackage[utf8]{inputenc}
\usepackage{graphicx}
\usepackage{cite}
\usepackage{multirow}
\begin{document}

\title{Few-Shot Class-Incremental Learning with Non-IID Decentralized Data}

\author{Cuiwei Liu, Siang Xu, Huaijun Qiu, and Jing Zhang, Zhi Liu, and Liang Zhao
\thanks{Cuiwei Liu, Siang Xu, Huaijun Qiu, Jing Zhang, and Liang Zhao are with the Department of Computer Science, the Shenyang Aerospace University, Shenyang, China. (E-mail: liucuiwei@sau.edu.cn; xusiang@stu.sau.edu.cn; 20220071@email.sau.edu.cn; 20230049@email.sau.edu.cn; lzhao@sau.edu.cn)}

\thanks{Zhi Liu is with the Department of Computer and Network Engineering, The University of Electro-Communications, Tokyo, Japan. (E-mail: liu@ieee.org)}

}



\maketitle

\begin{abstract}
Few-shot class-incremental learning is crucial for developing scalable and adaptive intelligent systems, as it enables models to acquire new classes with minimal annotated data while safeguarding the previously accumulated knowledge.
Nonetheless, existing methods deal with continuous data streams in a centralized manner, limiting their applicability in scenarios that prioritize data privacy and security.
To this end, this paper introduces federated few-shot class-incremental learning, a decentralized machine learning paradigm tailored to progressively learn new classes from scarce data distributed across multiple clients.
In this learning paradigm, clients locally update their models with new classes while preserving data privacy, and then transmit the model updates to a central server where they are aggregated globally.
However, this paradigm faces several issues, such as difficulties in few-shot learning, catastrophic forgetting, and data heterogeneity.
To address these challenges, we present a synthetic data-driven framework that leverages replay buffer data to maintain existing knowledge and facilitate the acquisition of new knowledge.
Within this framework, a noise-aware generative replay module is developed to fine-tune local models with a balance of new and replay data, while generating synthetic data of new classes to further expand the replay buffer for future tasks.
Furthermore, a class-specific weighted aggregation strategy is designed to tackle data heterogeneity by adaptively aggregating class-specific parameters based on local models performance on synthetic data.
This enables effective global model optimization without direct access to client data.
Comprehensive experiments across three widely-used datasets underscore the effectiveness and preeminence of the introduced framework.
\end{abstract}


\section{Introduction}
\IEEEPARstart{D}{eep} models have achieved remarkable success in various fields, attaining performance levels that are close to human capabilities.
However, we humans have the ability to quickly acquire new information from an ongoing series of tasks while retaining previously gained knowledge, which poses significant challenges to deep models.
In recent years, continual learning~\cite{wang2024comprehensive}, also referred to as lifelong or incremental learning, has garnered increasing attention.
Class-incremental learning (CIL)~\cite{masana2022class,9349197,zhou2023deep} as one of the most challenging scenarios, requires models to continuously learn new classes without the presence of task identification.

In real-world applications, CIL enables deep models to evolve for emerging classes, thereby significantly enhancing their flexibility and practicality.
Taking a fraud detection model in a banking system as an example, after it is deployed to multiple branches, novel types of fraud may emerge in various branches over time.
To promptly respond to these new threats, it is impractical to collect vast amounts of data.
In this scenario, the model must rapidly learn new fraud classes using limited data scattered across isolated branches, while maintaining its detection capabilities for previously observed fraud types.
This task is defined as Few-Shot Class-Incremental Learning (FSCIL) using decentralized data.

\begin{figure}[!t]
	\centering
	\includegraphics[width=\linewidth]{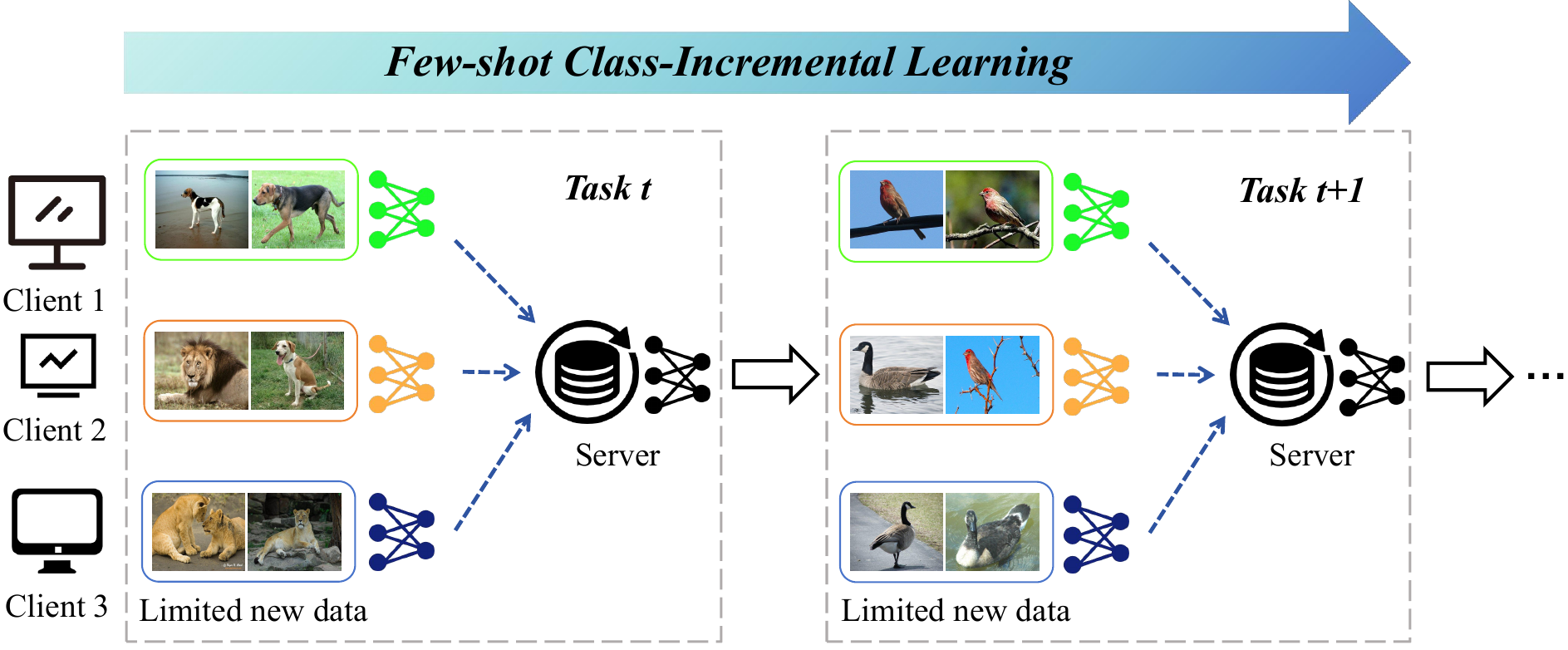}
	\caption{Illustration of the F2SCIL paradigm. In each incremental session, clients update their local models to gain new classes from scarce annotated data. The updated parameters are transmitted to a server where they are consolidated into a global model, which is then redistributed to clients for the subsequent task.}
	\label{fig:F2SCIL}
\end{figure}

Although significant progress has been made in FSCIL, most existing methods~\cite{tao2020few,zhang2021few,zhou2022forward,yang2023neural} deal with continuous data streams in a centralized manner, which is impractical for realistic scenarios that emphasize data privacy and security.
In the example of fraud detection in banking systems, continuous data streams from isolated branches cannot be shared or uploaded to a central server for model updates due to concerns about potential leakage or tampering of sensitive information.
Therefore, it is crucial to perform FSCIL using a decentralized approach, enabling multiple sites to evolve their models without sharing local data.


This work explores a decentralized machine learning paradigm referred to as Federated Few-Shot Class-Incremental Learning (F2SCIL). 
As depicted in Fig.~\ref{fig:F2SCIL}, clients (data owners) continually update their local models with scarce data to adapt to novel classes.
Subsequently, the updated parameters from local models are transmitted to a server for aggregation, producing a unified model that leverages distributed data across multiple clients while preserving data locality.
However, the F2SCIL paradigm faces several challenges.
Due to the scarcity of data accessible to each client (i.e., few-shot learning), the model encounters difficulties in learning new classes. 
Secondly, catastrophic forgetting is a significant issue as the model continually updates to incorporate new classes, potentially degrading its performance on previously learned classes. 
Thirdly, variations in data categories and sample sizes across clients, often described as non-Independent and Identically Distributed (Non-IID), result in data scarcity and heterogeneity.
These factors exacerbate catastrophic forgetting and pose substantial difficulties in designing an effective aggregation strategy for refining the global model.


This paper handles these challenges by presenting a novel Synthetic Data-Driven (SDD) framework. 
In the common CIL setting, data typically arrives in a stream, with each task or session introducing a new set of classes. 
Within the proposed SDD framework, each client updates its local model with newly acquired data to integrate new classes and utilizes synthetic data from the replay buffer to retain earlier knowledge.
Although some methods~\cite{rebuffi2017icarl,castro2018end,wu2019large} leverage the outputs of old models on replay data to guide the model learning, these knowledge distillation-based methods exhibit instability in FSCIL.
Drawing inspiration from~\cite{liu2022few}, we resort to re-labeling synthetic replay data with one-hot labels, subsequently incorporating these pseudo-labeled data into the current session's dataset.
However, a general consensus~\cite{zhang2022fine} in generating synthetic data is to include hard samples to enhance sample diversity and represent more class characteristics.
This also implies that relabeling synthetic data may introduce noisy labels, adversely affecting local model learning.
To address this, we develop a Noise-Aware Generative Replay (NAGR) module which fine-tunes local models using both new data and synthetic replay data to balance learning across old and new classes.
Subsequently, a conditional generator is constructed based on the ensemble of local models with the aim of producing synthetic data for the new classes in this session.
These synthetic data are further added to the replay buffer to assist the local model learning in future sessions.

Furthermore, a novel Class-Specific Weighted Aggregation (CSWA) strategy is devised to deal with the degradation of global model performance caused by data heterogeneity among clients.
This strategy calculates the weights of class-specific parameters in model aggregation by evaluating the performance of local models on each new class.
Notably, the original data cannot leave its owner in the F2SCIL paradigm. 
Therefore, we innovatively use the generated synthetic data in this session to simulate new class distributions, which facilitates the performance evaluation of local models and supports adaptive aggregation of the global model.

This paper makes the following key contributions:
\begin{itemize}
	\item
	We initiate the exploration of the F2SCIL paradigm, specifically targeting scenarios where a global model continuously learns new classes from limited data distributed across multiple clients.
	\item
	We propose a new framework called SDD, which utilizes synthetic data to memorize previously learned knowledge in client-side incremental learning and improve server-side model integration.
	\item
	We highlight the potential issue of noisy labels in synthetic data and propose a Noise-Aware Generative Replay (NAGR) module that alleviates catastrophic forgetting in a data-free manner.
	\item
	We point out that the data scarcity in F2SCIL makes the non-IID nature of data from different clients particularly challenging. 
	To this end, we devise a Class-Specific Weighted Aggregation (CSWA) strategy aimed at achieving efficient fusion of client models.
\end{itemize}

\section{related work}
\subsection{Few-shot class-incremental learning}
CIL allows models to acquire knowledge from an ongoing stream of training data while minimizing the loss of previous information.
Mainstream approaches fall into four categories.
First, replay-based methods~\cite{rebuffi2017icarl,rolnick2019experience,zhu2021prototype} incorporate data from earlier tasks while learning new ones to help maintain the model's memory.
Second, regularization-based methods~\cite{kirkpatrick2017overcoming,lopez2017gradient} modify the loss function by including regularization terms to steer the optimization direction.
Third, model-based methods~\cite{yan2021dynamically,wang2022foster,zhou2022model,wang2022learning,wang2022dualprompt} employ distinct model architectures or parameters for each task to prevent potential forgetting.
Lastly, algorithm-based methods~\cite{li2017learning,wu2019large} focus on designing algorithms that preserve the knowledge from earlier tasks.

FSCIL is more challenging since it requires the model to acquire new classes from limited annotated samples.
The TOPIC framework~\cite{tao2020few} is the first to introduce the concept of FSCIL, employing Neural Gas networks to map the spatial relationships within the feature space across various classes for knowledge encoding.
\cite{liu2022few} presents an entropy-regularized data-free replay method that learns a generator to produce synthetic data.
S3C~\cite{kalla2022s3c} addresses the FSCIL task through a stochastic classifier and a self-supervision approach. 
CEC~\cite{zhang2021few} integrates the topological structure of graph models with incremental models.
FACT~\cite{zhou2022forward} tackles this task from the perspective of forward compatibility, allocating multiple virtual prototypes in the feature space as a preservation space. 
ALICE~\cite{peng2022few} integrates an angular penalty loss to enhance feature clustering, training the backbone network with both base class data and synthetic data.
This work creates additional space for accommodating new classes and utilizes cosine similarity for classification.
NC-FSCIL~\cite{yang2023neural} proposes a neural collapse-based framework to reduce the divergence between features of old classes and well-trained classifiers.

Although existing FSCIL methods have shown promising results, they generally perform well under the assumption of centralized training data storage.
As a result, these methods may struggle to address challenges in distributed training scenarios, where the data from each incremental session comes from different clients.

\subsection{Federated continual learning}
Federated Learning~\cite{mcmahan2017communication} is a decentralized approach that allows multiple data sources to collaboratively develop a unified model.
Recently, it has achieved considerable success in various research and industry areas.
Federated Continual Learning (FCL) is a newly emerging topic that integrates the concepts of federated learning with continual (or incremental) learning.
In addition to the catastrophic forgetting issue on the client side, the FCL task also introduces new challenges, notably managing interference between clients and improving communication performance.
FedWeIT~\cite{yoon2021federated} tackles these issues by breaking down parameters into global, local, and task-adaptive components.
Each client selectively assimilates information from others through a weighted aggregation of task-adaptive parameters.
However, due to communication overhead, FedWeIT still requires a buffer for data rehearsal, which is deemed infeasible in our F2SCIL paradigm where samples are limited.
CFeD~\cite{ma2022continual} combines FCL with knowledge distillation, where the model obtained in the previous task acts as a teacher.
Additionally, a client-side distillation strategy is introduced to address the data heterogeneity problem. 
~\cite{hendryx2021federated} utilizes prototype networks, leveraging a server model that has been pre-trained in the initial task.
Nevertheless, unlike the proposed FCL setting, this work argues that new classes can overlap with old classes, which is not consistent with standard class-incremental learning. 
The TARGET framework~\cite{zhang2023target} demonstrates that non-IID data exacerbate catastrophic forgetting in federated learning and utilizes previously learned global model to train a generator, transferring knowledge from earlier tasks.
DCID~\cite{zhang2022deep} designs a decentralized incremental learning paradigm that requires to share some samples of clients for knowledge distillation.
MFCL~\cite{babakniya2024data} develops a data-free generative model learned at the server side to mitigate catastrophic forgetting and ensure data privacy.

However, these FCL methods rely on collecting and integrating abundant data from multiple sources and struggle to gain sufficient knowledge from limited data in our F2SCIL paradigm.
This data scarcity not only complicates the task of distinguishing new classes from existing ones but also exacerbates the model discrepancies caused by the heterogeneity of data across clients.

\subsection{Data-free replay method}
The data-free replay methods in CIL generally require training a generator model to produce replay data, which are used to retain previously learned knowledge.
DeepInversion~\cite{yin2020dreaming} introduces a model inversion technique that creates synthetic data for specific classes using random perturbations.
Some methods~\cite{chen2019data,micaelli2019zero} employ generative adversarial networks to synthesize data, utilizing the previously learned model as a discriminator while optimizing the generator with a teacher-student architecture.
Recently, another group of methods~\cite{smith2021always,xin2021memory} integrate the idea of generative replay with deep inversion to address this problem.
\cite{liu2022few} also attemptes to add entropy loss to the training of the generator to produce more uncertain replay samples in the FSCIL scenario.

In federated learning, FedGen~\cite{zhu2021data} develops a compact generator to enhance client-side training and consolidate the knowledge from local models without relying on previous data.
FedFTG~\cite{zhang2022fine} investigates the input space of local models using a server-side generator to produce synthetic data and carries out real-time fine-tuning of the global model through data-free knowledge distillation.
DENSE~\cite{zhang2022dense} is a data-free one-shot federated learning framework that applies data-free approaches to federated learning for model aggregation.

Unlike the previous methods, we train a conditional generator with an ensemble of local models to continuously produce synthetic data for new classes in the F2SCIL paradigm.
The synthetic data provide replay samples to mitigate catastrophic forgetting in client-side FSCIL and also serve as a validation set for assessing local models, facilitating more effective model aggregation.

\begin{figure*}[!t]
	\centering
	\includegraphics[width=\textwidth]{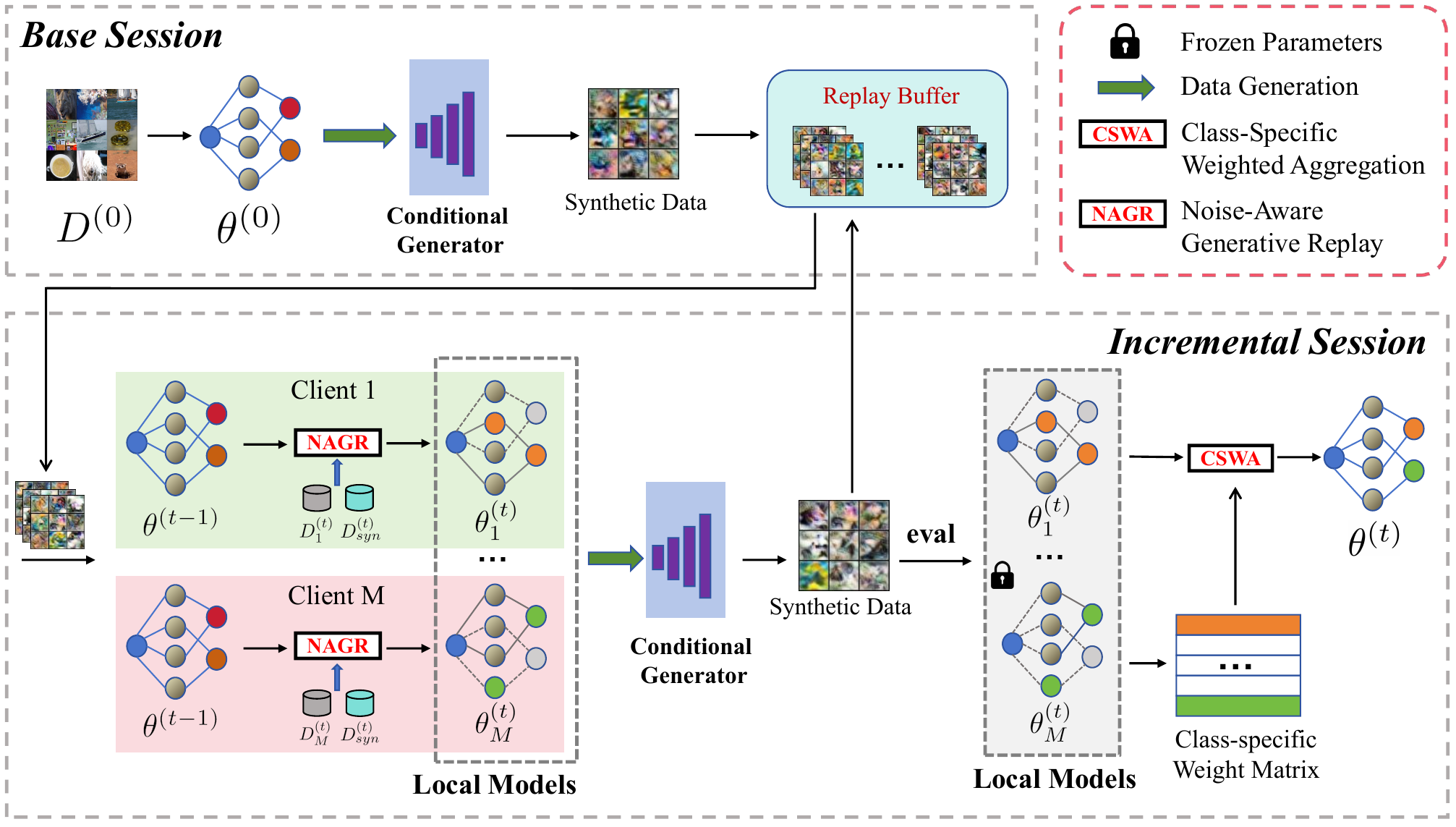}
	\vskip -5pt
	\caption{Illustration of the proposed SDD framework. In the base session, an initial model is trained with abundant data of base classes centrally. In an incremental session, clients employ the NAGR module to adapt to new classes. A conditional generator is trained based on client-side local models to produce synthetic data, which is then used to assess the performance of these models. This process creates a class-specific weight matrix for model aggregation. Continuously generated synthetic data are kept in a replay buffer to retain old knowledge for incremental learning in future sessions.}
	\label{fig:SDD}
	\vskip -10pt
\end{figure*}

\section{Method}
\subsection{Problem formulation}
First, we introduce the F2SCIL paradigm including a base session to learn data-abundant classes and $T$ incremental sessions for novel and data-scarce classes. 
Let $D = \{D^{(t)}\}_{t=0:T}$ represents the collection of training datasets corresponding to sessions indexed from $0$ to $T$.
Each $D^{(t)}$ is defined as $D^{(t)} = \{(x, y) \mid x \in X^{(t)}, y \in Y^{(t)}\}$, including examples from an image set $X^{(t)}$ annotated with labels taken from a fixed group of categories $Y^{(t)}$. 
Note that the training data for different sessions are mutually exclusive, i.e., $X^{(p)} \cap X^{(q)} = \varnothing$ and $Y^{(p)} \cap Y^{(q)} = \varnothing$ for $p \neq q$.
Furthermore, $D^{(0)}$ is a centralized dataset used in the base session to train an initial model $\theta ^ {(0)}$, while the data in each $D^{(1)}, D^{(2)}, \ldots, D^{(T)}$ is distributed across multiple clients for the subsequent incremental sessions.
The goal of the $t$-th incremental session is to evolve the model $\theta ^{(t-1)}$ into $\theta ^{(t)}$, utilizing the corresponding data $D^{(t)}$.

Specifically, during the $t$-th incremental session, the limited data in~$D^{(t)}$ is organized in the N-way K-shot format, where the label space $Y^{(t)}$ comprises $N$ distinct classes, each with a total of $K$ training images.
Notably, $D^{(t)}$ is distributed heterogeneously (i.e., non-IID) among $M$ clients in the F2SCIL paradigm. 
In simpler terms, $D^{(t)}$ comprises data sourced from $M$ clients, where the data belonging to the $m$-th client is denoted as $D^{(t)}_m$.

\subsection{The overview framework}

To address the challenges in F2SCIL as previously discussed, we propose a data-free framework called SDD.
The overview of SDD is depicted in~Fig.~\ref{fig:SDD}.
For the base session, we utilize the dataset $D^{(0)}$ to learn an initial model $\theta ^ {(0)}$ in a centralized manner.
Following the traditional supervised learning framework, $\theta ^ {(0)}$ is optimized by the cross-entropy loss function to ensure strong classification capabilities for base classes.
Next, the initial model is employed to provide supervisory signals for training a conditional generator, which can effectively capture the key feature distributions of the base classes and generate highly representative pseudo samples.
These generated samples are then added to a pre-established replay buffer, providing knowledge of previously learned classes for subsequent incremental sessions.

For incremental session $t$, the model $\theta ^{(t-1)}$ is updated to $\theta ^{(t)}$ through distributed learning on limited data of the new classes $D^{(t)}$.
First, the $m$-th client performs FSCIL to optimize its local model $\theta _m ^{(t)}$.
This process involves learning and adapting to the new classes using local data $D^{(t)}_m$.
To avoid catastrophic forgetting, a set of pseudo samples are drawn from the replay buffer to review and consolidate knowledge of the previously learned classes. 
Next, a conditional generator is constructed using the ensemble of local models, aiming to produce pseudo samples that represent the new classes in the current incremental session.
These newly generated pseudo samples are then employed as validation data to assess how well the local models perform on the new classes, resulting in a class-specific weight matrix to guide the aggregation of the global model $\theta ^{(t)}$.
At last, pseudo samples of new classes are further incorporated into the replay buffer, providing knowledge of these classes for subsequent incremental sessions.

\subsection{Few-shot class-incremental learning} \label{sec:NAGR}
On each client, our goal is to train a local model from the global model learned in the previous session. 
In incremental session $t$, the training of model ${\theta}^{(t)}_m$ at client $m$ can be defined as a local FSCIL problem, with the primary challenges of catastrophic forgetting and model overfitting.
We present an NAGR module that leverages synthetic replay data to retain previously learned knowledge and addresses the potential issue of noisy labels in synthetic data using a noise-robust loss.

Specifically, knowledge preservation for the old classes is achieved by using a set of synthetic data $D_{syn}^{(t)}$ drawn from the replay buffer.
As confirmed by~\cite{rebuffi2017icarl,li2017learning}, knowledge distillation is a typical and effective method in class-incremental learning, utilized to transfer knowledge from previous tasks through replay samples.
However, the recent work~\cite{liu2022few} points out that it is non-trivial to balance the classification loss for new samples with the distillation loss for replay samples in FSCIL scenarios.
So it suggests re-labeling the synthetic replay samples using one-hot encoding based on the old model and applying classification loss to these pseudo-labeled data.
This approach allows the synthetic dataset to be represented as $D_{syn}^{(t)}=\{(\widetilde{x},y^*)\}$, where $\widetilde{x}$ and $y^*$ indicate a synthetic sample and its pseudo-label.

However, we argue that there exist noisy labels in $D_{syn}^{(t)}$ due to the inclusion of hard samples in synthetic data generation to enhance sample diversity.
These noisy pseudo-labels can serve as misleading signals and negatively impact the local model learning.
Therefore, it is imperative to develop a mechanism that makes the learning process more robust to such noise, ensuring that the model can accurately capture the previously learned knowledge present in the synthetic replay data.

To this end, we introduce a noise-robust loss function~\cite{wang2019symmetric}, which optimizes the local model with the synthetic replay data of old classes by 
\begin{equation}
	\begin{aligned}
		&\mathcal{L}_{old}(\widetilde{x},y^*;{\theta}^{(t)}_m) =\alpha CE({\theta}^{(t)}_m(\widetilde{x}),y^*) + \beta RCE({\theta}^{(t)}_m(\widetilde{x}),y^*) \\
		&=\alpha\sum_{C_B}y^*log({\theta}^{(t)}_m(\widetilde{x}))+\beta\sum_{C_B}{\theta}^{(t)}_m(\widetilde{x})log(y^*),
	\end{aligned}
\end{equation}
where ${\theta}^{(t)}_m(\widetilde{x})$ indicates the probability distribution output from model ${\theta}^{(t)}_m$ given an input $\widetilde{x}$.
$C_B$ represents the number of previous classes, $\alpha$ and $\beta$ are hyper-parameters to balance the two items.
The combination of CE and RCE exhibits a symmetry.
The CE loss ensures the accurate prediction of true labels, thereby minimizing classification errors on clean data.
Meanwhile, RCE mitigates the influence of noisy labels by relying on the model's predicted distribution ${\theta}^{(t)}_m(\widetilde{x})$.
This approach encourages the model to focus on replay samples with high confidence, enhancing its robustness to noisy pseudo-labels.

For few samples of new classes, the model is trained using the standard cross-entropy loss.
This process is formulated by
\begin{equation}
	\label{eq:client}
	\mathcal{L}_{new}(x,y;{\theta}^{(t)}_m)=CE({\theta}^{(t)}_m(x),y),
\end{equation}
where $(x,y)$ is a sample data pair from the local dataset $D^{(t)}_m$ in the current incremental session.

Finally, the total objective function $\mathcal{L}_{client}$ for optimizing the local model $\theta^{(t)}_m$ can be formulated as a combination of $\mathcal{L}_{new}$ and $\mathcal{L}_{old}$.
\begin{equation}
	\mathcal{L}_{client} = \mathcal{L}_{new} + {k} \cdot \mathcal{L}_{old},
\end{equation}
where $k$ indicates the weight for the noise-robust loss applied to the synthetic data of old classes.

In addition, to address the issue of the model overfitting to the limited new data, we employ a fine-tuning strategy to update the local model.
Concretely, we fine-tune the parameters of the backbone and the old classifiers with a low learning rate, while training the parameters of the newly added classifiers with a higher learning rate.
This approach helps prevent excessive updates to the model parameters that could lead to overfitting the new classes.

\subsection{Data generation}
\begin{figure}[!t]
	\centering
	\includegraphics[width=\linewidth]{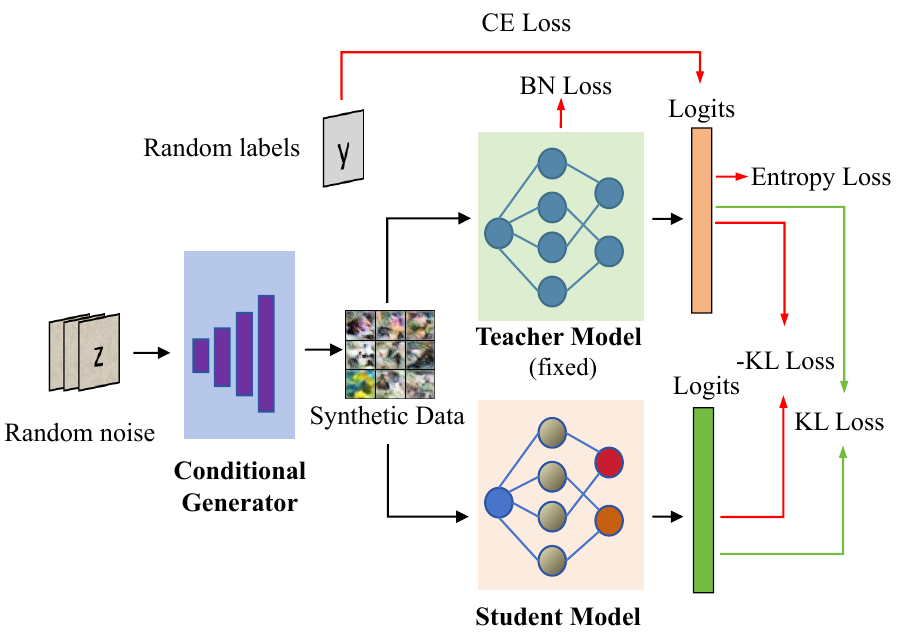}
	\vskip -5pt
	\caption{An illustration of training a conditional generator. We employ a teacher-student architecture to jointly optimize the conditional generator and student model through an adversarial learning mechanism, with guidance from the teacher model. The red and green lines denote the training loss for updating the generator and the student model, respectively.}
	\label{fig:generation}
\end{figure}

Given a well-trained model, our goal is to build a conditional generator $\theta_G$ for synthesizing pseudo-samples that mimic the original data distribution. 
The synthetic data are expected to satisfy the following key characteristics: fidelity, diversity, stability, and transferability. 
To achieve this, we adopt a teacher-student architecture where the condition generator and the student model are jointly optimized through an adversarial learning scheme, as depicted in Fig.~\ref{fig:generation}.
In the base session, the well-trained initial model functions as the teacher, while the ensemble of local models serves as the teacher in the subsequent incremental sessions.
Note that the teacher model only provides supervisory signals and does not update its parameters.
Following~\cite{zhang2023target}, we initialize a student model $\theta_S$ which transfers knowledge by emulating predictions of the teacher on synthetic samples, thereby learning similar decision boundaries.
The generator seeks to make knowledge transfer more challenging for the student model by generating synthetic samples near the decision boundaries.
Through this adversarial learning mechanism, the generator is able to produce diverse and challenging samples.
The training process consists of two iterative stages: 
\begin{itemize}
	\item Optimization of the generator: The generator is continuously optimized to synthesize diverse and hard pseudo-samples that conform to the distribution of the teacher.
	\item Optimization of the student model: The student model is trained on synthetic data to acquire knowledge from the teacher and assess the effectiveness of the synthetic data.
\end{itemize}

\subsubsection{\textbf{Optimization of the generator}}
Briefly, the generator $\theta_G$ seeks to produce a synthetic sample $\widetilde{x}=\theta_G (z,\widetilde{y})$ that aligns with the data distribution of the teacher model, where $z$ denotes a perturbation extracted from a standard Gaussian distribution and the label $\widetilde{y}$ is randomly chosen according to the predefined distribution $Y^{(t)}$.

We first obtain the logits $O(\widetilde{x})$, namely the outputs of the last fully connected layer in the teacher network.
This process can be formulated as
\begin{equation}
	O(\widetilde{x}) = \begin{cases}
		f^{t}(\widetilde{x};{\theta}^{(t)}), & \text{if } t = 0, \\
		\frac{1}{M}\sum_{m=1}^M f^{t}(\widetilde{x};{\theta}^{(t)}_m), & \text{if } t \geq 1.
	\end{cases}
\end{equation}
where ${\theta}^{(t)}$ denotes the global model and ${\theta}^{(t)}_m$ indicates local model at the $m$-th client in session $t$.
$M$ is the total number of clients.
The function $f^{t}(\cdot;\cdot)$ extracts outputs of the last fully connected layer, corresponding to the new classes in session $t$.
Particularly, the initial model ${\theta}^{(0)}$ learned in a centralized manner is employed as the teacher in the base session.
In the subsequent incremental sessions, we use the ensemble of local models rather than the global model as the teacher, because the generator is trained before aggregating the local models into the global model, as shown in Fig.~\ref{fig:SDD}.

\textbf{Fidelity.} We employ the standard cross-entropy loss to guarantee that synthetic replay samples are correctly categorized into a specific class $\widetilde{y}$ with high confidence.
\begin{equation}
	\mathcal{L}_{CE}(\widetilde{x},\widetilde{y};{\theta}_G)=CE(Softmax(O(\widetilde{x})),\widetilde{y}).
\end{equation}
By minimizing~$\mathcal{L}_{CE}$, the synthetic data can fit the distribution of class~$\widetilde{y}$.
In fact, during the training of the generator, relying only on CE loss often generates synthetic data that remain distant from the decision boundary defined by the teacher model, leading to the issue of overfitting.

\textbf{Diversity.}
Another objective of the generator is to ensure the diversity of the synthetic data.
To promote the diversity of generated synthetic data and incorporate more challenging examples, we increase the information entropy of the teacher model's predictions on these data.
Specifically, the information entropy of the probability distribution $p=[p_1,...,p_c]$ is computed as
\begin{equation}
	H_{info}(p)=-\frac{1}{c}\sum_{i=1}^{c}{p}_{i}\log({p}_{i}).
\end{equation}
The entropy loss of a synthetic sample is formulated as
\begin{equation}
	\mathcal{L}_{En}(\widetilde{x};{\theta}_G)=-H_{info}(Softmax(O(\widetilde{x}))).
\end{equation}

\textbf{Stability.} Early studies~\cite{yin2020dreaming,zhang2022dense} have demonstrated that the Batch Normalization (BN) statistics is beneficial for the convergence of generator.
BN layers play a crucial role in reducing internal covariate shifts, a phenomenon where the distribution of network activations changes as training data forward propagate through the layers of a neural network.
By ensuring that the activations of synthetic data align with these precomputed statistics of teacher models, the generator produces samples that closely resemble the distribution of the original dataset.
The BN statistic constraint is given by
\begin{equation}
	\begin{aligned}
		\mathcal{L}_{BN}(\widetilde{x};\boldsymbol{\theta}_G)=
		\frac{1}{M}\sum_{m=1}^M\sum_{l}\left(\|\mu_l(\widetilde{x})-\mu_{m,l}\|+\right.\\
		\left.\left\|\sigma_l^2(\widetilde{x})-\sigma_{m,l}^2\right\|\right),
	\end{aligned}
\end{equation}
where~$\mu_l(\widetilde{x})$ and $\mu_{m,l}$ represent the mean of activations at the $l$-th batch normalization layer in the generator and the $m$-th teacher model, respectively. Similarly, $\sigma_l^2(\widetilde{x})$ and $\sigma_{m,l}^2$ denote the variance of activations at the same layers in the generator and teacher model, respectively.

\begin{figure}[t]
	\centering
	\includegraphics[width=\linewidth]{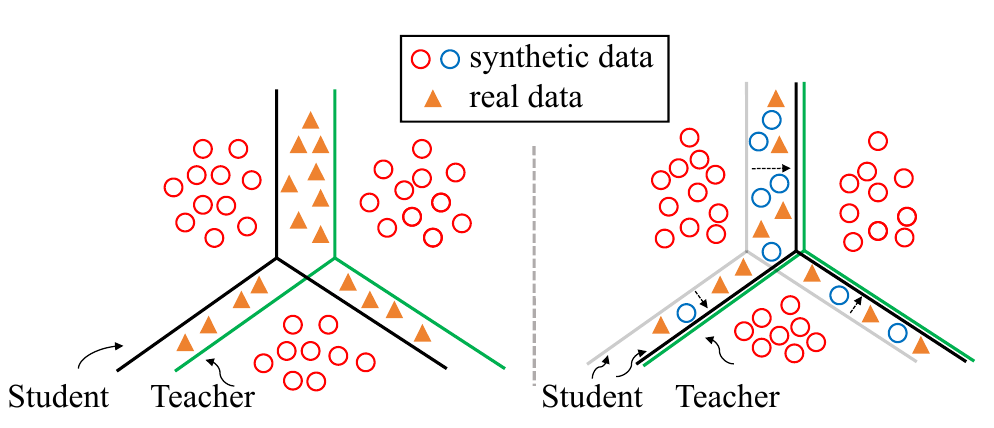}
	\caption{An illustration of the synthetic data and decision boundaries of both teacher and student models. Left Panel: Without the loss $\mathcal{L}_{KL}$, the generated synthetic data (red circles) are far from the teacher's decision boundary. Right Panel: With $\mathcal{L}_{KL}$, the generator produces more challenging synthetic data (blue circles) closer to the decision boundary, aiding the student in better mimicking the teacher's decision boundary.}
	\label{fig:distribution}
\end{figure}

\textbf{Transferability.}
The core motivation for employing a teacher-student architecture is to drive the generator to produce hard samples near the decision boundaries.
The student mimics the teacher's outputs on synthetic samples to gain valuable knowledge, while the generator aims to produce samples that lie on different sides of the decision boundaries of the student and teacher, as illustrated in Fig.~\ref{fig:distribution}.
To achieve this, we encourage synthetic samples to induce prediction discrepancies between the teacher and student models by maximizing the Kullback-Leibler divergence between their output logits.
\begin{equation}
	\mathcal{L}_{KL}(\widetilde{x};{\theta}_G)=- \omega KL\left(O(\widetilde{x}),O_s(\widetilde{x};{\theta}_S)\right),
\end{equation}
where $O(\widetilde{x})$ and $O_s(\widetilde{x};{\theta}_S)$ denote the logits produced by the teacher and the student, respectively.
Particularly, $\omega=1$ if the student and the teacher predict different classes for the sample $\widetilde{x}$; otherwise, $\omega=0$.

\textbf{The objective function} of the generator is composed of the above loss functions and can be formulated as
\begin{equation}
	\begin{aligned}
		\label{eq:G}
		\mathcal{L}_{G}(\widetilde{x},\widetilde{y};\boldsymbol{\theta}_G)&=  
		\lambda_1\cdot\mathcal{L}_{CE}(\widetilde{x},\widetilde{y};{\theta}_G) 
		+\lambda_2\cdot\mathcal{L}_{En}(\widetilde{x};{\theta}_G) \\
		&\quad+\lambda_3\cdot\mathcal{L}_{BN}(\widetilde{x};{\theta}_G)
		+\lambda_4\cdot\mathcal{L}_{KL}(\widetilde{x};{\theta}_G),
	\end{aligned}
\end{equation}
where the pre-defined parameters $\lambda_1$, $\lambda_2$, $\lambda_3$, and $\lambda_4$ are utilized to weight the four items.

\subsubsection{\textbf{Optimization of the student model}}
We propose leveraging a student model to assist in training the generator.
Given a synthetic sample, the student model tries to mimic outputs of the teacher model through optimization of the Kullback-Leibler divergence
\begin{equation}
	\label{eq:S}
	\mathcal{L}_{S}(\widetilde{x};{\theta}_S)=KL\left(O(\widetilde{x}),O_s(\widetilde{x};{\theta}_S)\right),
\end{equation}
so that it can learn similar decision boundaries.
In contrast, the generator strives to generate samples close to the category decision boundary, making the knowledge transfer of the student model difficult.
Adversarial learning against the generator facilitates the training of a high-performing student model and helps to collect a diverse and challenging set of synthetic samples.

\subsection{Model aggregation}

Most existing model aggregation algorithms in federated learning adopt model-wise weighted averaging to derive the global model from local ones.
However, they do not consider the differences between the internal parameters of various local models.
In the proposed SDD framework, we develop a CSWA strategy to achieve efficient fusion of local models by leveraging synthetic data generated in the current session.
This strategy accounts for the differences in local model parameters caused by the scarce and non-IID data in FSCIL and implements parameter-wise weighted aggregation for new classes based on the class-specific weight matrix.

In incremental session $t$, parameters of the local model $\theta_m^{(t)}$ are divided into old parameters $W_{\mathrm{old}}^{t,m}$ inherited from the previous session and new parameters $W_{\mathrm{new}}^{t,m}$ added for classifying new classes in the current session.
We only fine-tune the parameters $W_{\mathrm{old}}^{t,m}$ to better preserve knowledge of old classes. Therefore, desired results can be achieved by averaging these old parameters $W_{\mathrm{old}}^{t,m}$ on the server side.
This process is formulated by
\begin{equation}
	\label{eq:w_old}
	\widetilde{W}_{\mathrm{old}}^{t}=\sum_{m=1}^M\frac{N_m^{(t)}}{N^{(t)}}{{W}_{\mathrm{old}}^{t,m}},
\end{equation}
where $\widetilde{W}_{\mathrm{old}}^{t}$ represents the consolidated outcomes of old parameters in the global model.
$N_m^{(t)}$ signifies the count of training samples at client $m$, whereas $N^{(t)}$ denotes the aggregate number of training samples across all clients in the current session $t$.

Parameters ${{W}_{\mathrm{new~}}^{t,m}} \in \mathbb{R}^{L \times c}$ introduced in the current session for classifying new classes are formulated as
\begin{equation}
	W_{\mathrm{new}}^{t,m}=\begin{bmatrix}w_{1,1}^m&\cdots&w_{1,i}^m&\cdots&w_{1,c}^m\\\vdots&\ddots&&\vdots&\vdots\\w_{l,1}^m&&w_{l,i}^i&&w_{l,c}^m\\\vdots&&&\ddots&\vdots\\w_{L,1}^m&\cdots&w_{L,i}^m&\cdots&w_{L,c}^m\end{bmatrix},
\end{equation}
where $c$ and $L$ signify the count of new classes and the feature dimension, respectively.
${{W}_{\mathrm{new~}}^{t,m}}$ are randomly initialized at the beginning of the incremental session and subsequently updated with data of new classes in the FSCIL process.
During model updates on the client side, these newly added parameters are likely to overfit the limited training data.
Therefore, we employ the synthetic data $\widetilde{X}_t$ derived from the generator in the current session $t$ to evaluate the performance of multiple local models on new classes.
Specifically, we send the synthetic data $\widetilde{X}_t$ to the local model on client $m$ and statistically estimate its accuracy on new classes, expressed as 
\begin{equation}
	A_{\mathrm{new}}^{t,m}=\begin{bmatrix}a_m^1,\ldots,a_m^i,\ldots,a_m^c\end{bmatrix},
\end{equation}
where $a_m^i$ indicates the accuracy of local model on client $m$ for class $i$.
Then the class-specific weight matrix $A_{\mathrm{new}}^t \in \mathbb{R}^{M \times c}$ of all local models can be expressed as
\begin{equation}
	\label{eq:A_new}
	A_{\mathrm{new}}^t=\begin{bmatrix}A_{\mathrm{new}}^{t,1};A_{\mathrm{new}}^{t,2};...;A_{\mathrm{new}}^{t,M} \end{bmatrix}.
\end{equation}

Then, the accuracy vector $\{A_{\mathrm{new}}^{t,m}\}_{m=1:M}$ are taken as the weights of local models in the aggregation of new parameters.
A high accuracy $a_m^i$ indicates that the local model on client $m$ provides reliable classification for class $i$ in this session, so parameters pertaining to class $i$ in this model should be assigned high confidence in model aggregation.
Specifically, we perform element-wise multiplication between the new parameters $W_{\mathrm{new}}^{t,m}$ of a local model and the corresponding accuracy vector $A_{\mathrm{new}}^{t,m}$, producing a weighted parameter matrix ${\widetilde{W}_{\mathrm{new}}^{t,m}}$ by
\begin{equation}
	\label{eq:m_new}
	\begin{aligned}
		{\widetilde{W}_{\mathrm{new}}^{t,m}}&=A_{\mathrm{new~}}^{t,m}\textstyle\bigotimes{W_{\mathrm{new}}^{t,m}}\\
		&=\begin{bmatrix}a_m^1 w_{1,1}^m&\cdots&a_m^i w_{1,i}^m&\cdots&a_m^c w_{1,c}^m\\\vdots&\ddots&&\vdots&\vdots\\a_m^1 w_{l,1}^m&&a_m^i w_{l,i}^m&&a_m^c w_{l,c}^m\\\vdots&&&\ddots&\vdots\\a_m^1 w_{L,1}^m&\cdots&a_m^i w_{L,i}^m&\cdots&a_m^c w_{L,c}^m\end{bmatrix}.
	\end{aligned}
\end{equation}
Next, we sum the parameters of the newly updated classifiers from each client and obtain the aggregation result for this portion of the parameters by
\begin{equation}
	\label{eq:w_new}
	\widetilde{W}_{\mathrm{new}}^{t}= \sum_{m=1:M} {\widetilde{W}_{\mathrm{new}}^{t,m}}.
\end{equation}
Finally, the aggregated new parameters $\widetilde{W}_{\mathrm{new}}^t$ are combined with the aggregated old parameters $\widetilde{W}_{\mathrm{old}}^t$ to form the global model $\theta^{(t)}$ of incremental session $t$:
\begin{equation}
	\label{eq:total}
	\theta^{(t)}\leftarrow({\widetilde{W}_{\mathrm{old}}^t}, {\widetilde{W}_{\mathrm{new}}^t}).
\end{equation}

\begin{algorithm}[!t]
	\renewcommand{\algorithmicrequire}{\textbf{Input:}}
	\renewcommand{\algorithmicensure}{\textbf{Output:}}
	\caption{Baseline approach.}
	\label{alg:baseline}
	\begin{algorithmic}[1]
		\REQUIRE Distributed training set $\{D_m^{(t)}\}_{m=1:M}$ in incremental session $t$, synthetic data $D_{syn}^{(t)}$ drawn from the replay buffer, and the global model $\theta^{(t-1)}$ of the preceding session. 
		\ENSURE The refined global model $\theta^{(t)}$.
		\STATE Initialize the conditional generator $\theta_G$ and the student model $\theta_S$.
		\STATE Initialize the newly generated synthetic dataset in this session: $\widetilde{X}_t = \varnothing$.
		\STATE Distribute model $\theta^{(t-1)}$ to clients as the initial model $\theta_m^{(t)}$.
		
		\STATE\textbf{for} each round $r = 1,2,..., \textbf{to }  R $ \textbf{do}
		\STATE\hspace{1em}\textbf{for} each client $m = 1,2,..., \textbf{to }  M$
		\STATE\hspace{1em}\textbf{in parallel do}
		\STATE\hspace{2em}\textbf{for} $(x,y) \in D_m^{(t)}$ and $\widetilde{x} \in D_{syn}^{(t)}$ \textbf{do}
		\STATE\hspace{3em}Update local model $\theta_m^{(t)}$ by minimizing\\
		\hspace{3em} $\mathcal{L}=CE(\theta_m^{(t)}(x),y)+ k \cdot KL(\theta^{t-1}(\widetilde{x}),\theta_m^{(t)}(\widetilde{x}))$.
		\STATE\hspace{2em}\textbf{end for}
		\STATE\hspace{1em}\textbf{end for}
		\STATE\hspace{1em} \textit{// Data generation}
		\STATE\hspace{1em}\textbf{for} each iteration $i = 1,2,..., \textbf{to } K$ \textbf{do}
		\STATE\hspace{2em}Draw a batch of random perturbations and categories $\{(z_i,\widetilde{y}_i)\}_{i=1}^b$;
		\STATE\hspace{2em}Synthesize $\{\widetilde{x}_i\}_{i=1}^b$ with  $\{(z_i,\widetilde{y}_i)\}_{i=1}^b$ and $\theta_G$;
		\STATE\hspace{2em}Update $\theta_G$ by minimizing $\mathcal{L}_{G}$ in~(\ref{eq:G});
		\STATE\hspace{2em}Update $\theta_S$ by minimizing $\mathcal{L}_{S}$ in~(\ref{eq:S});
		\STATE\hspace{2em}Add a batch of data into $\widetilde{X}_t$.
		\STATE\hspace{1em}\textbf{end for}
		\STATE\hspace{1em}Aggregate local models into the global model:\\
		\STATE\hspace{1em}$\theta^{(t)}\leftarrow\sum_{m=1}^M\frac{N_m^{(t)}}{N^{(t)}}\theta_m^{(t)}$.
		\STATE\textbf{end for}
		\STATE\hspace{0em}Add the newly generated synthetic dataset $\widetilde{X}_t$ into the replay buffer.
	\end{algorithmic}
\end{algorithm}

\begin{algorithm}[!t]
	\renewcommand{\algorithmicrequire}{\textbf{Input:}}
	\renewcommand{\algorithmicensure}{\textbf{Output:}}
	\caption{The proposed SDD.}
	\label{alg:SDD}
	\begin{algorithmic}[1]
		\REQUIRE Distributed training set $\{D_m^{(t)}\}_{m=1:M}$ in incremental session $t$, synthetic data $D_{syn}^{(t)}$ drawn from the replay buffer, and the global model $\theta^{(t-1)}$ of the preceding session. 
		
		\ENSURE The refined global model $\theta^{(t)}$.
		
		\STATE Initialize the conditional generator $\theta_G$ and the student model $\theta_S$.
		\STATE Initialize the newly generated synthetic dataset in this session: $\widetilde{X}_t = \varnothing$.
		\STATE Distribute model $\theta^{(t-1)}$ to clients as the initial model $\theta_m^{(t)}$.
		
		\STATE\textbf{for} each round $r = 1,2,..., \textbf{to }  R $ \textbf{do}
		
		\STATE\hspace{1em}\textbf{for} each client $m = 1,2,..., \textbf{to }  M$ 
		\STATE\hspace{1em}\textbf{in parallel do}
		\STATE\hspace{2em}Update local model $\theta_m^{(t)}$ by minimizing $\mathcal{L}_{client}$ in~(\ref{eq:client}).
		\STATE\hspace{1em}\textbf{end for}
		\STATE\hspace{1em} \textit{// Data generation}
		\STATE\hspace{1em}\textbf{for} each iteration $i = 1,2,..., \textbf{to } K$ \textbf{do}
		\STATE\hspace{2em}Draw a batch of random perturbations and categories $\{(z_i,\widetilde{y}_i)\}_{i=1}^b$;
		\STATE\hspace{2em}Synthesize $\{\widetilde{x}_i\}_{i=1}^b$ with  $\{(z_i,\widetilde{y}_i)\}_{i=1}^b$ and $\theta_G$;
		\STATE\hspace{2em}Update $\theta_G$ by minimizing $\mathcal{L}_{G}$ in~(\ref{eq:G});
		\STATE\hspace{2em}Update $\theta_S$ by minimizing $\mathcal{L}_{S}$ in~(\ref{eq:S});
		\STATE\hspace{2em}Add a batch of data into $\widetilde{X}_t$.
		\STATE\hspace{1em}\textbf{end for}
		\STATE\hspace{1em} \textit{// Model aggregation}
		\STATE\hspace{1em}Calculate aggregated old parameters $\widetilde{W}_{\mathrm{old}}^t$ using~(\ref{eq:w_old}).
		\STATE\hspace{1em}\textbf{for} each client $m = 1,2,..., \textbf{to }  M$ 
		\STATE\hspace{1em}\textbf{in parallel do}
		\STATE\hspace{2em}Calculate $A_{\mathrm{new}}^{t,m}$ by evaluating local model $\theta_m^{(t)}$ on $\widetilde{X}_t$;
		\STATE\hspace{2em}Calculate new parameters ${\widetilde{W}_{\mathrm{new}}^{t,m}}$ by~(\ref{eq:m_new}).
		\STATE\hspace{1em}\textbf{end for}
		\STATE\hspace{1em}Calculate aggregated new parameters $\widetilde{W}_{\mathrm{new}}^t$ by~(\ref{eq:w_new});
		\STATE\hspace{1em}Integrate $\widetilde{W}_{\mathrm{old}}^t$ and $\widetilde{W}_{\mathrm{new}}^t$ into the global model $\theta^{(t)}$ \\
		\STATE\hspace{1em}by~(\ref{eq:total}).
		\STATE\textbf{end for }
		\STATE\hspace{0em}Add the newly generated synthetic dataset $\widetilde{X}_t$ into the replay buffer.
	\end{algorithmic}
\end{algorithm}

\subsection{Baseline approach}
This paper introduces a new learning paradigm F2SCIL, necessitating the proposal of a baseline approach for reference.
The baseline approach employs a classic knowledge distillation algorithm from class-incremental learning to update the local models on the client side and adopts FedAvg~\cite{mcmahan2017communication} as the model aggregation algorithm.
Details of this approach are described in Algorithm~\ref{alg:baseline}.
Taking the $t$-th incremental session as an example, we first disseminate the global model from the preceding session to clients as their initial models.
Subsequently, a client adapts its local model to new classes with the limited acquired data and employs knowledge distillation with synthetic samples of old classes drawn from the replay buffer to retain previously learned knowledge.
At last, these locally evolved models are aggregated into a new global model on the server side, using the FedAvg strategy.
This model is conveyed back to clients as the initial model in the next session.

Building on the baseline approach, the proposed SDD framework addresses several key issues in F2SCIL.
In the update of the client-side local models, the NAGR module is introduced to manage the catastrophic forgetting problem.
In the model aggregation phase, the CSWA strategy is developed to consider the significant differences in local model parameters caused by non-IID data, thereby ensuring efficient and effective model aggregation.
The overall learning procedure of SDD for an incremental session is outlined in Algorithm~\ref{alg:SDD}.

\section{Experiments}

\subsection{Datasets}
Comprehensive experiments are performed across three datasets to verify the feasibility of the proposed SDD.
\subsubsection{\textbf{CIFAR100}}
This is a popular benchmark in incremental learning and federated learning, designed for image classification tasks.
It comprises a total of 60,000 images across 100 classes, each of which contains 500 training samples and 100 testing samples.
All images are 32x32 pixels in size.

\subsubsection{\textbf{miniImageNet}}
This dataset contains 100 classes, each furnished with 500 images designated for training purposes and 100 images reserved for testing.
These classes are selected from the larger ImageNet dataset~\cite{russakovsky2015imagenet}.
Each image is 84x84 pixels in size, resized from the original ImageNet images.

\subsubsection{\textbf{TinyImageNet}} 
This dataset is also a reduced version of the ImageNet dataset~\cite{russakovsky2015imagenet}.
It contains 200 classes, with 500 training images and 50 testing images per class.
Each image is 64x64 pixels, resized from the original higher-resolution images in ImageNet.

\renewcommand{\arraystretch}{1.2}
\begin{table*}[t]
	\centering
	\caption{Classification accuracy (\%) of different methods under the F2SCIL paradigm on the CIFAR100 dataset.}
	\label{tab:CIFAR100}
	\begin{tabular}{c|ccccccccc|c|c}
		\hline
		\multirow{2}{*}{Methods} & \multicolumn{9}{c|}{Sessions} & Average & \multirow{2}{*}{Improvement}\\
		\cline{2-10}
		& 0 & 1 & 2 & 3 & 4 & 5 & 6 & 7 & 8 & accuracy & \\
		\hline
		Finetune + FL & 74.53 & 57.85 & 44.04 & 42.73 & 41.94 & 34.39 & 30.52 & 25.98 & 23.76 & 41.75 & +20.02\\
		iCaRL\cite{rebuffi2017icarl} + FL & 74.53 & 68.18 & 61.40 & 51.91 & 47.22 & 42.15 & 34.26 & 30.17 & 31.00 & 48.98 & +12.79 \\
		EWC\cite{kirkpatrick2017overcoming} + FL & 74.53 & 63.68 & 54.43 & 49.51 & 44.12 & 41.00 & 38.50 & 36.40 & 31.04 & 48.13 & +13.64 \\
		LwF\cite{li2017learning} + FL & 74.53 & 66.03 & 55.03 & 47.23 & 42.10 & 36.95 & 34.82 & 27.87 & 24.62 & 45.46 & +16.31\\
		ERDFR\cite{liu2022few} + FL & 74.53 & 67.94 & 62.77 & 58.33 & 54.41 & 50.60 & 47.48 & 45.05 & 42.56 & 55.96 & +5.81\\
		Target\cite{zhang2023target} & 74.53 & 66.91 & 59.17 & 52.88 & 50.52 & 47.15 & 43.77 & 39.21 & 38.95 & 52.57 & +9.20\\
		MFCL\cite{babakniya2024data} & 73.87 & 67.69 & 62.51 &57.27 & 53.70 & 50.24 & 47.61 & 45.10 & 42.23 & 55.58 & +6.19\\
		\hline
		Baseline  & 74.53 & 70.20 & 65.54 & 60.95 & 57.78 & 54.81 & 52.59 & 50.28 & 47.98 & 59.41 & +2.36\\
		Baseline w/ NAGR & 74.53 & 70.51 & 66.57 & 62.67 & 59.41 & 56.65 & 54.90 & 52.80 & 50.22 & 60.92 & +0.85\\
		Baseline w/ CSWA & 74.53 & 70.40 & 66.31 & 62.03 & 58.79 & 56.12 & 54.51 & 52.16 & 49.42 & 60.47 & +1.30 \\
		SDD (Ours) & 74.53 & \textbf{70.86} & \textbf{67.27} &\textbf{63.15} &\textbf{60.39} &\textbf{57.87} & \textbf{56.43} & \textbf{54.32} & \textbf{51.08} & \textbf{61.77} & - \\
		\hline
	\end{tabular}
\end{table*}

\begin{table*}[t]
	\centering
	\caption{Classification accuracy (\%) of different methods under the F2SCIL paradigm on the miniImageNet dataset.}
	\label{tab:miniImageNet}
	\vskip -5pt
	\begin{tabular}{c|ccccccccc|c|c}
		\hline
		\multirow{2}{*}{Methods} & \multicolumn{9}{c|}{Sessions} & Average  & \multirow{2}{*}{Improvement}\\
		\cline{2-10}
		& 0 & 1 & 2 & 3 & 4 & 5 & 6 & 7 & 8 & accuracy & \\
		\hline
		Finetune + FL & 71.28 & 64.03	&56.13	&51.51	&46.21	&35.38&	32.04	&25.59&	22.23 &44.93& +12.53\\
		iCaRL\cite{rebuffi2017icarl} + FL & 71.28&  61.12	& 57.83	& 51.59 & 45.91 & 40.24 & 36.63	& 34.79 & 30.01	& 47.71 &  +9.75 \\
		EWC\cite{kirkpatrick2017overcoming} + FL & 71.28 & 64.55	&55.19&	50.49	&46.91&	37.93&	37.93&	35.39	&29.18&	47.65 & +9.81\\
		LwF\cite{li2017learning} + FL &71.28 & 62.43	&55.83&	49.53&	45.05&	36.99&	31.63&	27.79	&24.88&	45.05 & +12.41\\
		ERDFR\cite{liu2022few} + FL & 71.28 & 64.65 & 59.00 & 54.12 & 49.65 & 45.81 & 42.89 & 40.20 & 37.70 & 51.70 & +5.76\\
		Target\cite{zhang2023target} & 71.28 & 62.63 & 56.17 & 51.36 & 47.60 & 44.34 & 42.91 & 41.04 & 38.99 & 50.70 & +6.76\\
		MFCL\cite{babakniya2024data} & 71.65 & 64.66 & 59.60 &54.96 & 50.76 & 46.41 & 42.47 & 40.00 & 37.68 & 52.02 & +5.44\\
		\hline
		Baseline  & 71.28 & 65.95&	61.04&	56.92&	53.38&	50.35	&47.57&	44.98&	42.84 & 54.92 & +2.54\\
		Baseline w/ NAGR &71.28 & \textbf{66.72}&	62.44	&58.77&	55.95&	53.07&	50.28&	47.63	&46.02 & 56.91 & +0.55\\
		Baseline w/ CSWA & 71.28 & 66.42	&62.26&	58.40	&55.21&	51.99&	49.19&	46.92	&45.27 & 56.33 & +1.13 \\
		SDD (Ours) & 71.28 & 66.32&\textbf{62.73}	&\textbf{59.61}	&\textbf{56.91}	&\textbf{54.09}&	\textbf{50.92}&	\textbf{48.56}&	\textbf{46.72}&\textbf{57.46} & - \\
		\hline
	\end{tabular}
\end{table*}

\subsection{Implementation details}
For the CIFAR100 and miniImageNet datasets, we utilize the first 60 classes to intensively train the initial model in the base session.
The leftover 40 classes are then partitioned into 8 incremental sessions for decentralized learning.
Each session introduces 5 novel classes, with each class comprising a total of 25 training samples.
For the TinyImageNet dataset, comprising 200 classes, the first 100 classes are employed in the base session, while the left 100 classes are incrementally introduced over the course of 10 subsequent sessions.
Each of these sessions incorporates 10 new classes, each containing 50 training samples.
Notably, the few-shot training samples within an incremental session are non-IID distributed among M clients, with the Dirichlet parameter set to 1.
In our implementations, $M$ is set to 5 across three datasets.

We employ distinct backbone networks and experimental settings on the three datasets.
ResNet20~\cite{he2016deep} serves as the backbone for the CIFAR100 dataset, while ResNet18~\cite{he2016deep} is used for both the miniImageNet and TinyImageNet datasets.
In the centralized base session, the initial model is trained for 100 epochs with a learning rate of 0.1, which is decayed by 0.1 at epochs 60 and 70, respectively.
In subsequent decentralized incremental sessions, the learning rate for both the backbone and the old classifier is set to 0.0001, while the learning rate for the newly introduced classifier is set to 0.1.
The model is optimized for 15, 30, and 40 epochs on the CIFAR100, miniImageNet, and TinyImageNet datasets, respectively.
In all of the incremental sessions, the communication round $R$ is fixed at 1.

When training the client-side local models on the CIFAR100 dataset, we fix the scaling factors $\lambda_1$, $\lambda_2$, $\lambda_3$, and $\lambda_4$, as well as hyper-parameters $\alpha$, $\beta$, and $k$, all to 1.
For the miniImageNet and TinyImageNet datasets, $\lambda_1$, $\lambda_2$, $\lambda_3$, and $\lambda_4$ are set to 10, 0.1, 1, and 1, respectively, while $\alpha$, $\beta$, and $k$ are set to 1, 1, and 0.5, respectively.

We iteratively train the generator $\theta_G$ and the student model $\theta_S$ for 100 epochs.
The generator is trained for 50 rounds per epoch, using the Adam optimizer with a learning rate of 0.001.
For training the student model, we use the SGD optimizer with a learning rate of 0.2 and a momentum of 0.9.

\subsection{Experimental results}
The results of Non-IID experiments on CIFAR100, miniImageNet, and TinyImageNet datasets are detailed in Tables~\ref{tab:CIFAR100}, \ref{tab:miniImageNet}, and \ref{tab:tinyImageNet}, respectively.
Session 0 represents the centralized learning in the base session, while sessions 1-8 represent the decentralized learning in incremental sessions. 
To evaluate the proposed SDD framework, we use the standard FSCIL metrics, including the overall accuracy of learned classes in a session, the average accuracy across all sessions, and the performance improvement of SDD over other methods.

The proposed SDD framework is compared with the baseline method and two variants, each incorporating either the NAGR module or the CSWA strategy separately. 
A fine-tuning experiment is conducted by training SDD solely on the training data in the current session.
We also include three representative CIL methods: data replay-based iCaRL~\cite{rebuffi2017icarl}, parameter-based EWC~\cite{kirkpatrick2017overcoming}, and distillation-based LwF~\cite{li2017learning}.
A data-free FSCIL method, ERDFR~\cite{liu2022few}, is implemented for a fair comparison in the few-shot learning setting.
The above methods are adapted to fit the experimental environment, where ``+FL" indicates the introduction of federated learning.
To be more specific, the training data in each incremental session are distributed across multiple clients for local training, and the global model is derived by calculating the average of parameters of local models.
Additionally, the comparison also includes two data-free federated continual learning methods: Target~\cite{zhang2023target} and MFCL~\cite{babakniya2024data}.
All experiments on the same dataset use a unified backbone network, with MFCL~\cite{babakniya2024data} making slight adjustments to the classification layer to adapt to its training strategy.

According to the experimental results depicted in Tables~\ref{tab:CIFAR100},~\ref{tab:miniImageNet},~\ref{tab:tinyImageNet}, we make the following observations.
\begin{itemize}
	\item
	Results on the three datasets indicate that fine-tuning with limited new data leads to a significant performance drop due to catastrophic forgetting.
	The data scarcity of new classes causes CIL methods~\cite{rebuffi2017icarl,kirkpatrick2017overcoming,li2017learning} and federated continual learning methods~\cite{zhang2023target,babakniya2024data} to be prone to overfitting, resulting in inferior overall performance.
	In contrast, the FSCIL method ERDFR~\cite{liu2022few} achieves relatively high and stable performance.
	\item
	Our baseline method outperforms all existing methods, showing improvements of approximately 3.45\%, 3.22\%, and 1.03\% over the best previous method, ERDFR~\cite{liu2022few}, on the CIFAR100, miniImageNet, and TinyImageNet datasets, respectively.
	This indicates the superiority of the baseline method for the challenging F2SCIL paradigm.
	\item 
	Further performance improvements are observed when each of the proposed modules is added to the baseline individually.
	The best results are achieved when both modules are used, as demonstrated by our SDD method.
	The SDD achieves significant improvements of 2.36\%, 2.54\% and 2.31\% over the baseline on the CIFAR100, miniImageNet, and TinyImageNet datasets, respectively.
\end{itemize}

\begin{table*}[t]
	\centering
	\caption{Classification accuracy (\%) of different methods under the F2SCIL paradigm on the tinyImageNet dataset.}
	\label{tab:tinyImageNet}
	\begin{tabular}{c|ccccccccccc|c|c}
		\hline
		\multirow{2}{*}{Methods} & \multicolumn{11}{c|}{Sessions} & Average  & \multirow{2}{*}{Improvement}\\
		\cline{2-12}
		& 0 & 1 & 2 & 3 & 4 & 5 & 6 & 7 & 8 & 9 & 10 & accuracy & \\
		\hline
		Finetune + FL & 64.64 & 57.33 & 51.32 & 45.45 & 39.64 & 37.69 & 35.39 & 31.41 & 30.56 & 27.17 & 19.74 & 40.03 & +8.42 \\
		iCaRL\cite{rebuffi2017icarl} + FL & 64.64 & 57.20	& 50.47	& 45.40 & 41.39 & 38.45 & 36.00	& 33.52 & 32.03	& 30.32 & 28.69 & 41.65 & +6.80 \\
		EWC\cite{kirkpatrick2017overcoming} + FL & 64.64 & 56.31 & 50.72 & 43.78 & 40.57 & 36.75 & 35.00 & 33.99 & 31.09 & 27.16 & 24.73 & 40.43 & +8.02\\
		LwF\cite{li2017learning} + FL & 64.64 & 57.80	& 51.83 & 45.35 & 40.40 & 37.52 & 35.80 & 29.79	&27.43&	25.04 & 21.54 & 39.74& +8.71\\
		ERDFR\cite{liu2022few} + FL & 64.64 & 58.93 & 54.28 &49.45 & 45.77 & 42.72 &40.01 & 37.73 &36.01 & 34.12 & 32.53 & 45.11& +3.34\\
		Target\cite{zhang2023target} & 64.64 & 57.56  & 52.45 & 48.06  & 44.21  & 40.83  & 38.02 &  35.76 & 33.40  & 31.62 & 30.17 & 43.34 & +5.11 \\
		MFCL\cite{babakniya2024data} & 64.62 & 58.31 & 52.87 & 47.69 & 43.76 & 39.79 & 36.67 & 32.24 & 30.04 & 27.71 & 24.93 &41.69 & +6.76\\
		\hline
		Baseline  & 64.64 & 59.13 &	55.05 &	50.42 &	47.39 &	44.25 & 41.61 &	39.51 &	37.00 & 35.26 & 33.29 &46.14 & +2.31\\
		Baseline w/ NAGR & 64.64 & 59.65 &	55.58	&51.75&	48.06&	45.32&	42.90&	40.67	&38.44 & 36.42 & 34.15 & 47.05 & +1.40\\
		Baseline w/ CSWA & 64.64 & 59.82	&55.37&	51.55	&47.93&45.29&	42.86&	40.21	&38.23 & 36.25 & 34.32 &46.95 &  +1.50 \\
		SDD (Ours) & 64.64 &\textbf{60.42} & \textbf{56.60}	&\textbf{53.03}	&\textbf{49.86}	&\textbf{46.95}&	\textbf{44.80}&	\textbf{42.18}&	\textbf{40.17}&\textbf{38.21} & \textbf{36.12} &\textbf{48.45} & - \\
		\hline
	\end{tabular}
\end{table*}

\subsection{Ablation studies}

\subsubsection{\textbf{Analysis of different loss functions}}
As mentioned in Section~\ref{sec:NAGR}, traditional replay-based class-incremental learning methods employ knowledge distillation loss to retain old knowledge and avoid catastrophic forgetting.
Recent work~\cite{liu2022few} addresses FSCIL problem by relabeling the synthetic replay data and optimizing the local models on both new data and replay data using the cross-entropy loss.
Considering that noise pseudo-labels might be generated during the relabeling process, the proposed NAGR module introduces a noise-robust loss to better retain old knowledge.

In order to assess the efficacy of the introduced NAGR module, we apply knowledge distillation loss, cross-entropy loss, and noise-robust loss to the synthetic replay data in the baseline approach.
Table~\ref{tab:loss} summarizes the final accuracy in the last incremental session and the average accuracy across all sessions on the CIFAR100 and miniImageNet datasets.
We can see that applying cross-entropy loss to relabeled replay data yields better performance than using knowledge distillation loss.
The noise-robust loss in the proposed NAGR module further enhances both the final accuracy and average accuracy compared to cross-entropy loss, demonstrating its effectiveness in handling noisy pseudo-labels of some replay samples.

\begin{table}[t]
	\centering
	\caption{Performance (\%) of applying different loss functions to the synthetic replay data in FSCIL.}
	\label{tab:loss}
	\begin{tabular}{c|cc|cc}
		\toprule
		\multirow{3}{*}{Loss}& \multicolumn{2}{c|}{CIFAR100} & \multicolumn{2}{c}{miniImageNet} \\
		\cline{2-5}
		& Final  &Average  &Final &Average \\
		& acc & acc & acc & acc \\
		\midrule
		Knowledge distillation loss& 47.98 & 59.41 & 42.84 & 54.92\\
		Cross-entropy loss& 48.77 & 59.75 & 44.05 & 55.58\\
		\textbf{Noise-robust loss} & \textbf{50.22} & \textbf{60.92} & \textbf{46.02} & \textbf{56.91}\\
		\bottomrule
	\end{tabular}
\end{table}

\begin{table}[t]
	\centering
	\caption{Performance (\%) of different model aggregation strategies. }
	\label{tab:aggregation}
	\begin{tabular}{c|cc|cc}
		\toprule
		\multirow{2}{*}{Method}& \multicolumn{2}{c|}{CIFAR100} & \multicolumn{2}{c}{miniImageNet} \\
		\cline{2-5}
		& Final acc &Average acc &Final acc &Average acc \\
		\midrule
		FedAvg & 47.98 & 59.41 & 42.84 & 54.92\\
		FedProx & 48.02 & 59.37 & 42.91 & 54.92\\
		FedMax & 47.45 & 58.89 & 41.42 & 53.88\\
		\textbf{CSWA} & \textbf{49.42} & \textbf{60.47} & \textbf{45.27} & \textbf{56.33}\\
		\bottomrule
	\end{tabular}
\end{table}

\subsubsection{\textbf{Analysis of different model aggregation strategies}}
Next, we assess the effectiveness of the proposed CSWA strategy in the F2SCIL paradigm by comparing it with classic model aggregation strategies, including FedAvg~\cite{mcmahan2017communication}, FedProx~\cite{li2020federated}, and FedMax~\cite{chen2021fedmax}.
Based on the baseline method, experiments are conducted on the CIFAR100 and miniImageNet datasets using different model aggregation strategies and the results are presented in Table~\ref{tab:aggregation}.
By analyzing the final accuracy achieved in the last session as well as the average accuracy of all sessions, we can conclude that our CSWA strategy outperforms the other strategies on both datasets.

\begin{figure}[t]
	\centering
	\subfloat[Final accuracy]{
		\includegraphics[width=0.47\linewidth]{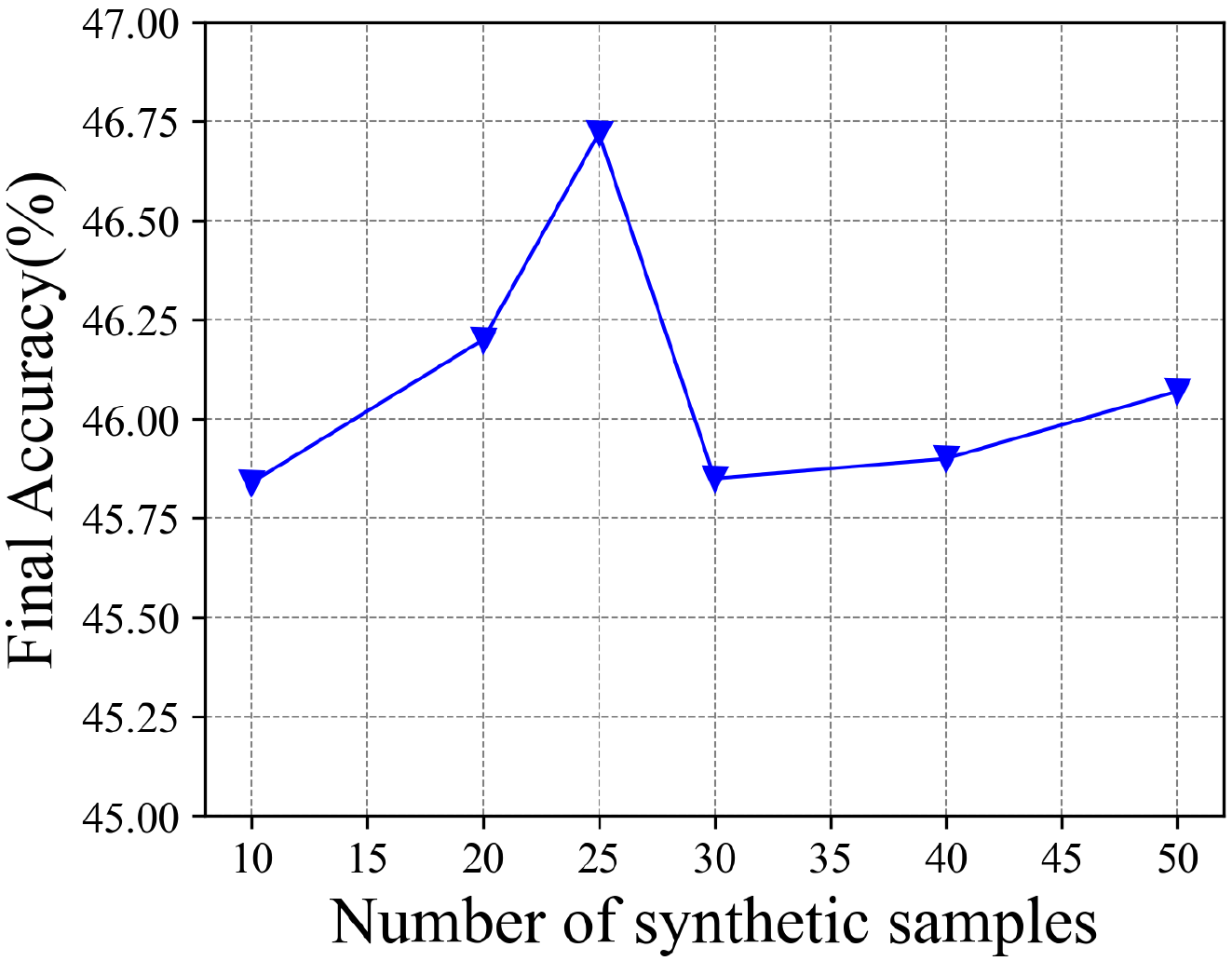}
		\label{fig.3(a)}
	}
	\hfill
	\subfloat[Average accuracy]{
		\includegraphics[width=0.47\linewidth]{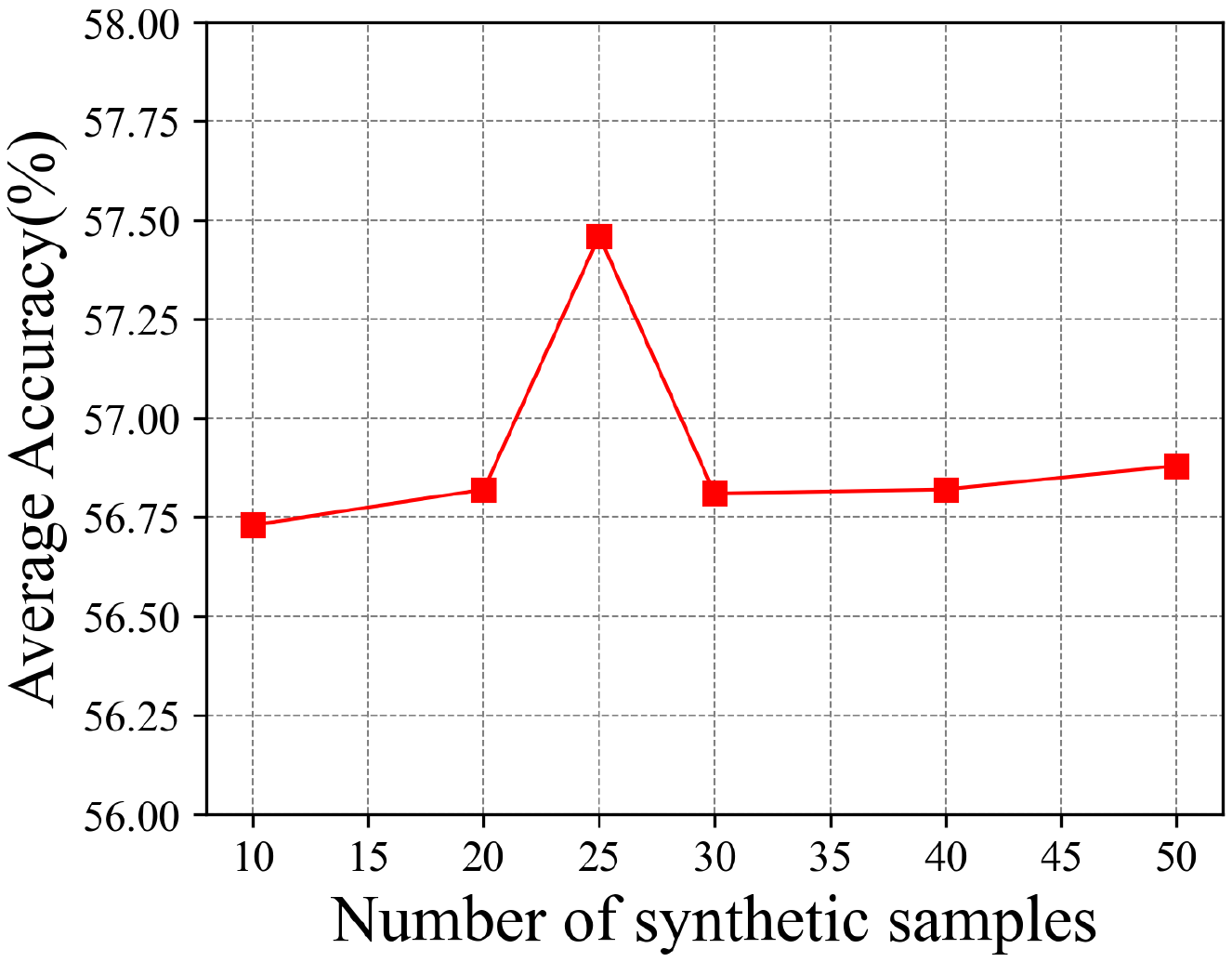}
		\label{fig.3(b)}
	}
	\caption{Performance of the proposed SDD with different numbers of synthetic replay samples on miniImageNet.}
	\label{fig:number}
\end{figure}

\subsubsection{\textbf{Analysis of the number of replay data}}
Fig.~\ref{fig:number} illustrates the effect of varying the number of synthetic replay samples.
In our setting, 125 training samples are randomly distributed among 5 clients in each incremental session, meaning each client has an average of 25 training samples for new classes.
The experimental results indicate that the model performs best when the amount of replay data is close to that of new training samples.
Using either fewer or more replay samples may decrease the overall performance, primarily due to the negative effects of data imbalance.
Indeed, limited replay data can be insufficient to retain old knowledge, while an excessive use of replayed data could potentially affect the adaptation to new classes.
Notably, we employ dynamic replay data across different batches and new synthetic data are drawn from the replay buffer for each batch training.

\begin{figure}[t]
	\centering
	\subfloat[Session 1]{
		\includegraphics[width=0.47\linewidth]{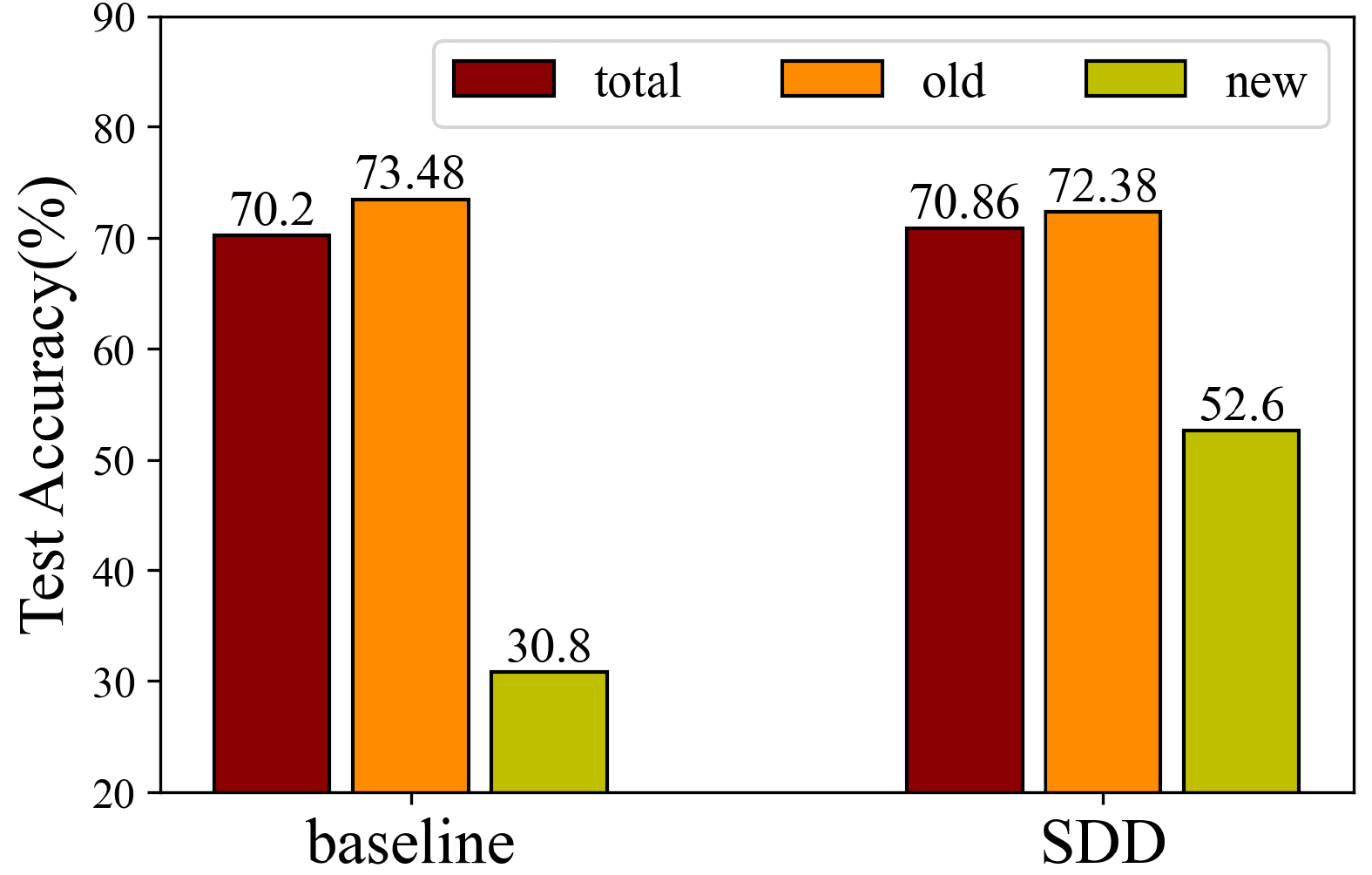}
		\label{fig:trade-off(a)}
	}
	\hfill
	\subfloat[Session 3]{
		\includegraphics[width=0.47\linewidth]{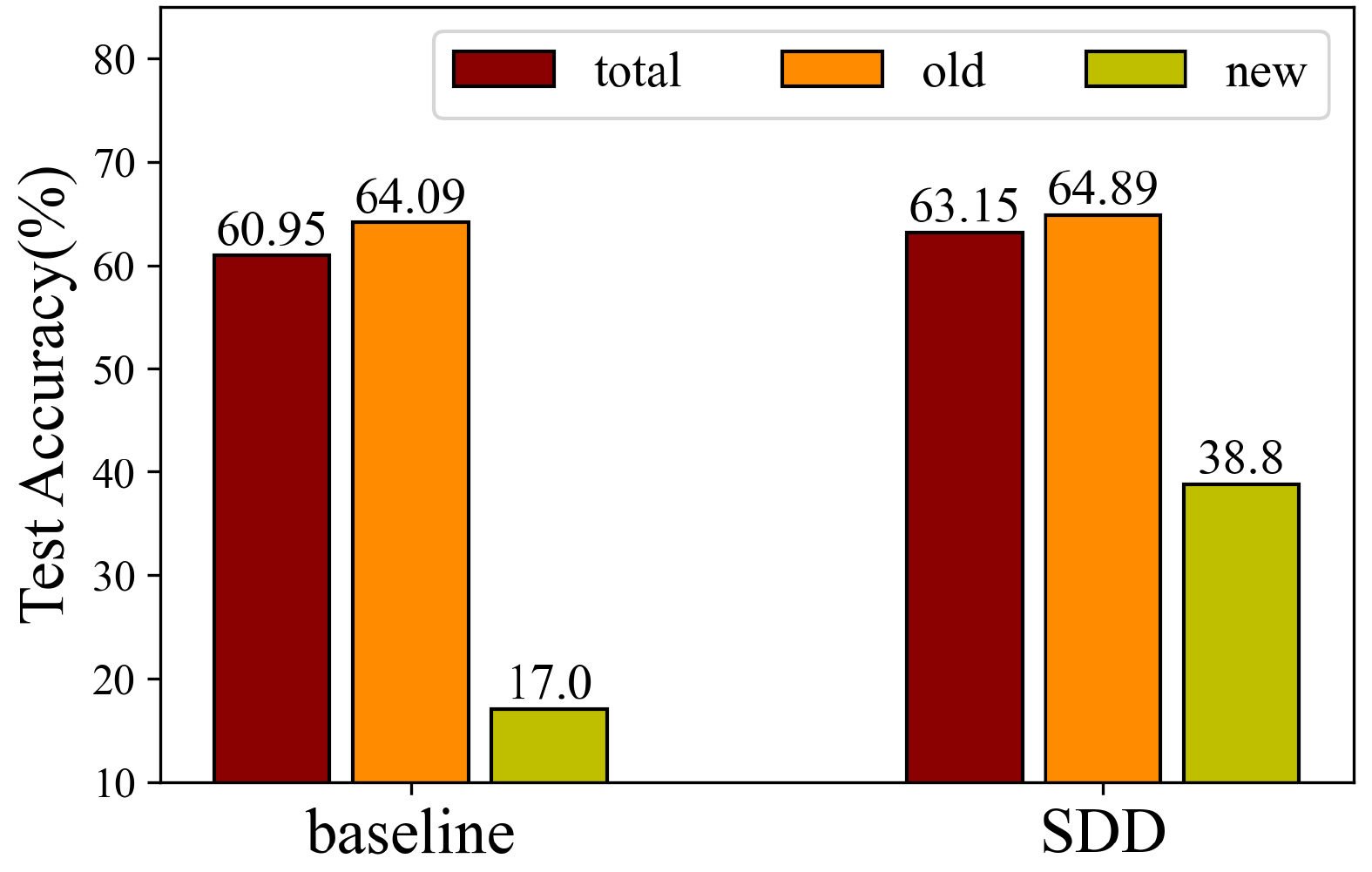}
		\label{fig:trade-off(b)}
	}
	\hfill
	\subfloat[Session 5]{
		\includegraphics[width=0.47\linewidth]{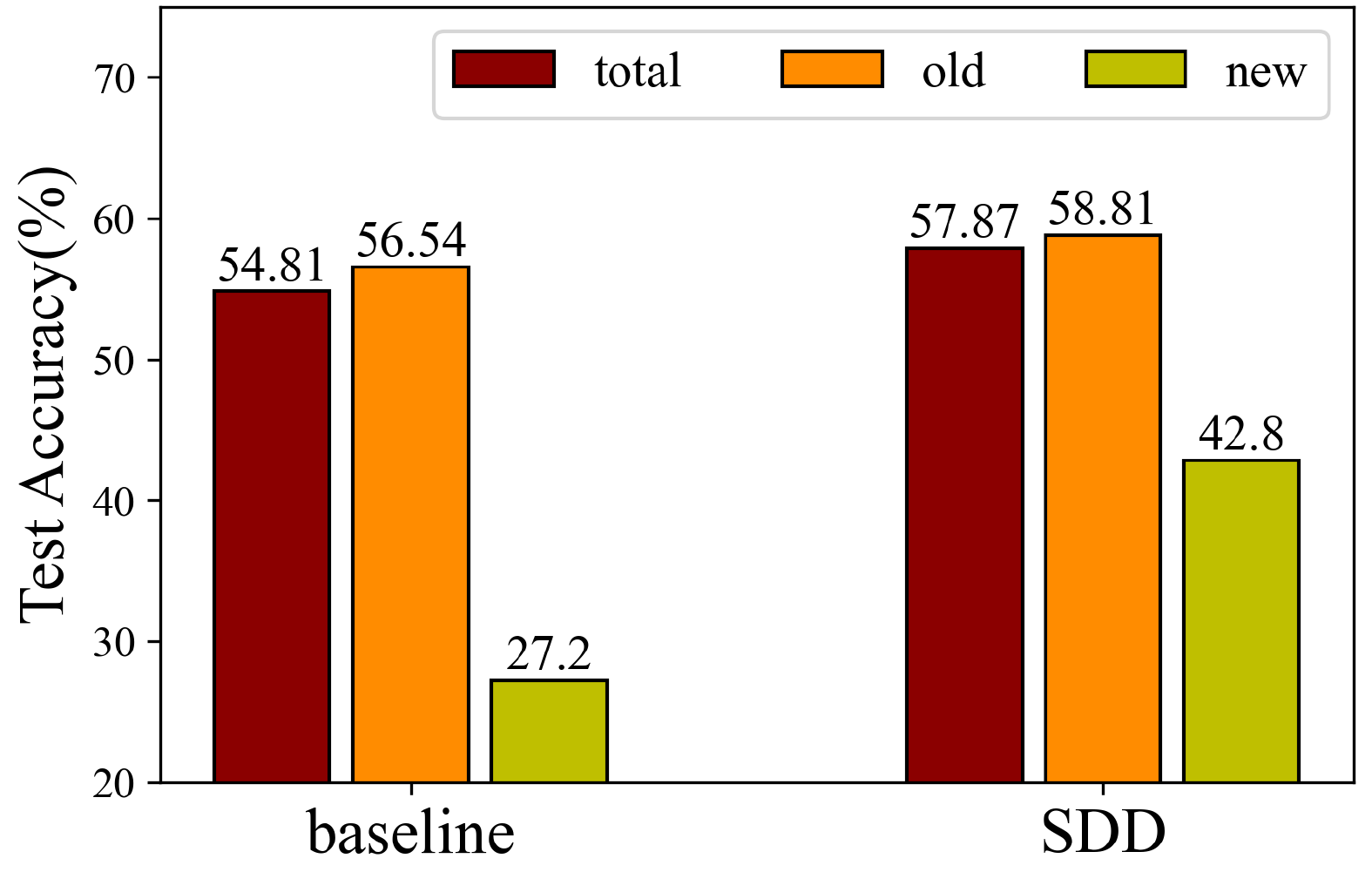}
		\label{fig:trade-off(c)}
	}
	\hfill
	\subfloat[Session 8]{
		\includegraphics[width=0.47\linewidth]{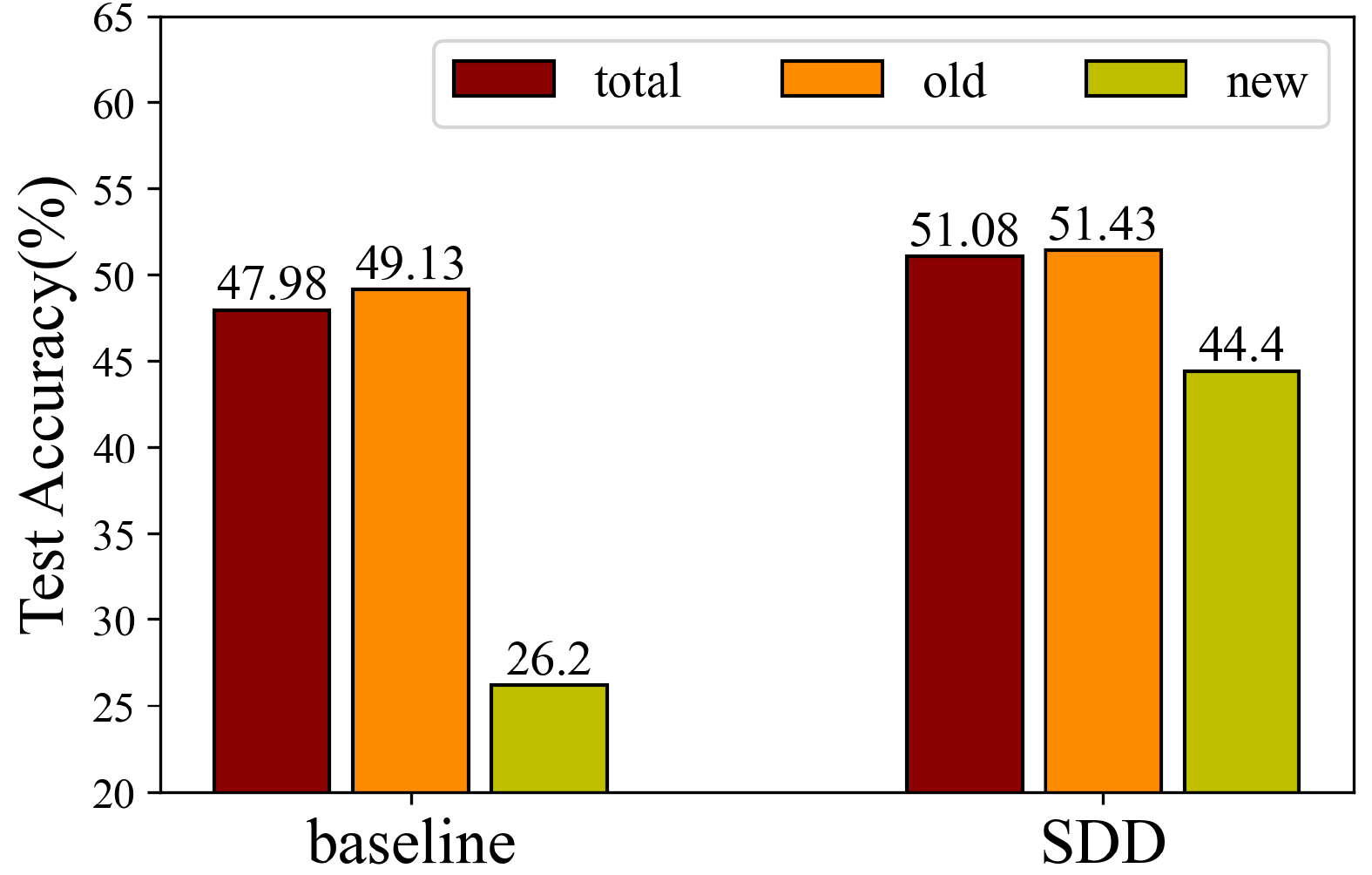}
		\label{fig:trade-off(d)}
	}
	\caption{Performance of the proposed SDD and the baseline for new and old classes across different incremental sessions on CIFAR100.}
	\label{fig:trade-off}
\end{figure}  

\subsubsection{\textbf{Analysis of the stability-plasticity trade-off}}
In the F2SCIL paradigm, base classes often dominate the dataset, leading to high overall accuracy even if performance on newly added classes is poor.
To evaluate the ability to learn new classes while retaining old knowledge, Fig.~\ref{fig:trade-off} presents the classification accuracy of the proposed SDD and the baseline for old and new classes across different incremental sessions.
In the first incremental session, SDD shows a 1.1\% decrease in accuracy for the 60 old classes compared to the baseline, but achieves a 21.8\% increase in accuracy for new classes.
In the final incremental session (Session 8), SDD demonstrates a 2.3\% improvement in accuracy for the 95 previously learned classes and an 18.2\% increase for new classes compared to the baseline. 
Similar trends are observed in other incremental sessions, indicating that the proposed SDD effectively balances stability for old classes with plasticity for new classes.



\begin{figure}[t]
	\centering
	\par 
	\subfloat[$\alpha=1$]{
		\includegraphics[width=0.3\linewidth]{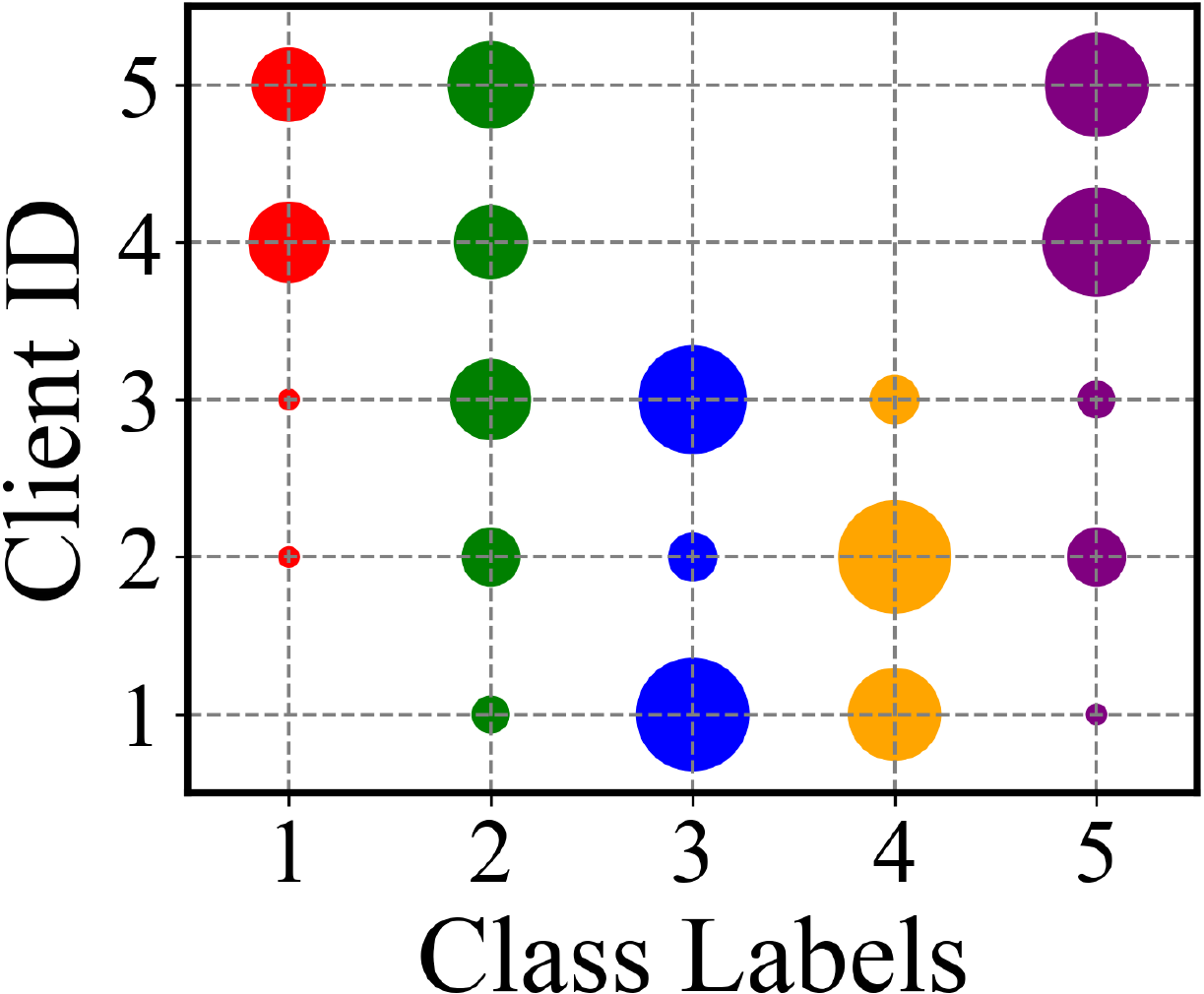}
		\label{fig.6(a)}
	}%
	\hfill
	\subfloat[$\alpha=0.5$]{
		\includegraphics[width=0.3\linewidth]{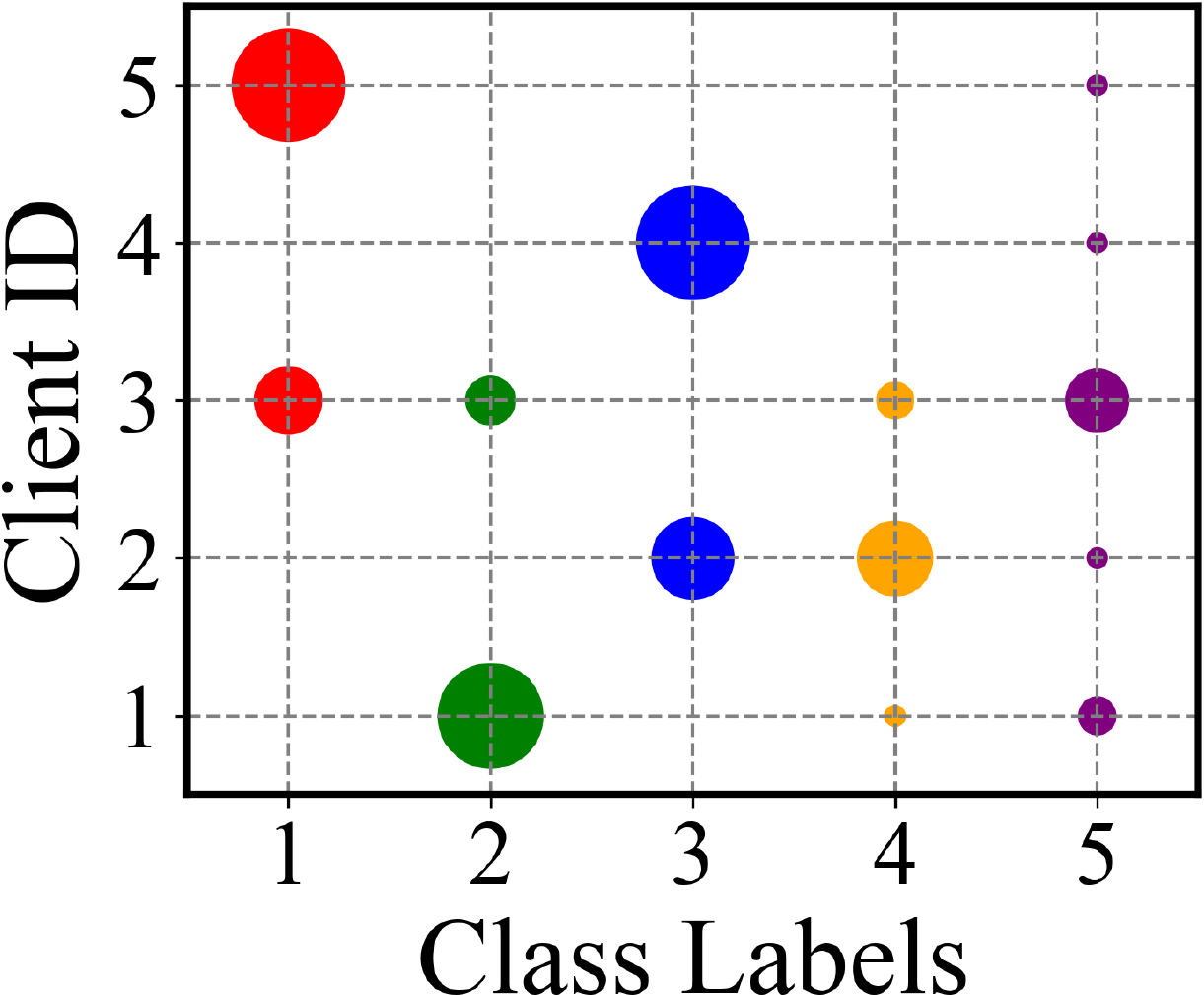}
		\label{fig.6(b)}
	}%
	\hfill
	\subfloat[$\alpha=0.1$]{
		\includegraphics[width=0.3\linewidth]{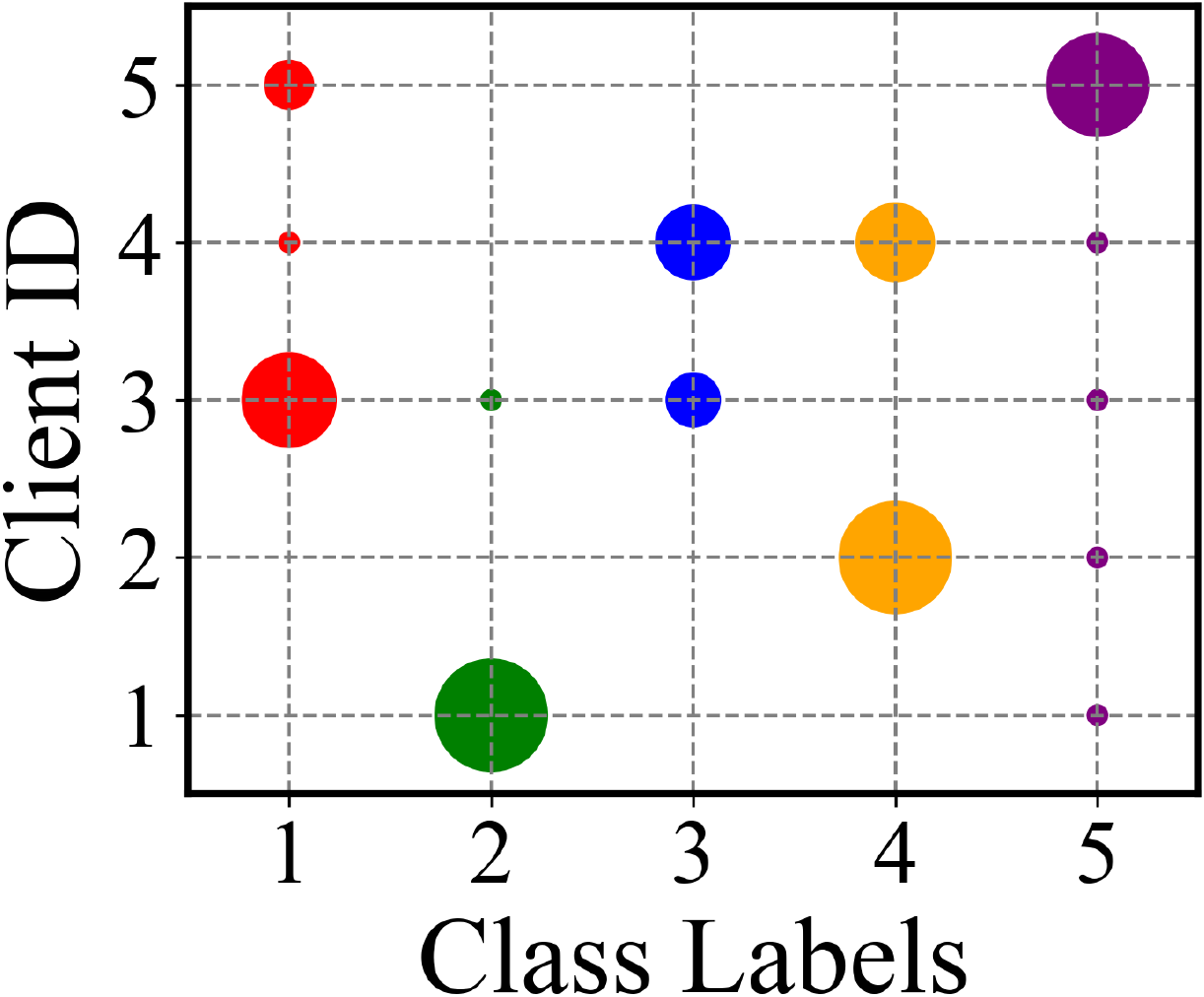}
		\label{fig.6(c)}
	}%
	\caption{Visualization of data distribution for each client with different Dirichlet distribution value $\alpha$. The larger the circle, the more training data of that class is present on the client.}
	\label{distributions}
\end{figure}

\begin{figure}[t]
	\centering
	\subfloat[Final accuracy]{
		\includegraphics[width=0.47\linewidth]{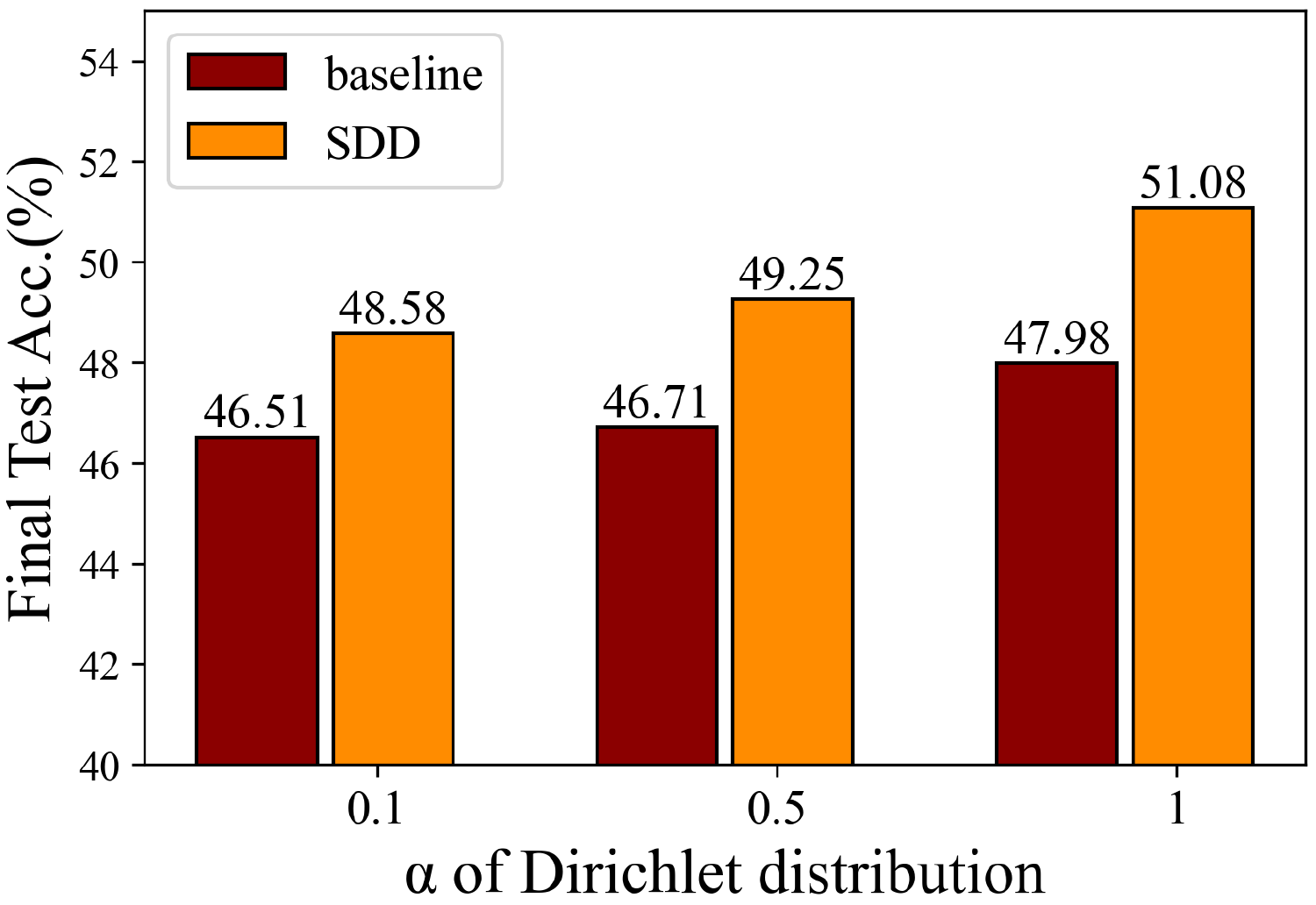}
		\label{fig.7(a)}
	}
	\hfill
	\subfloat[Average accuracy]{
		\includegraphics[width=0.47\linewidth]{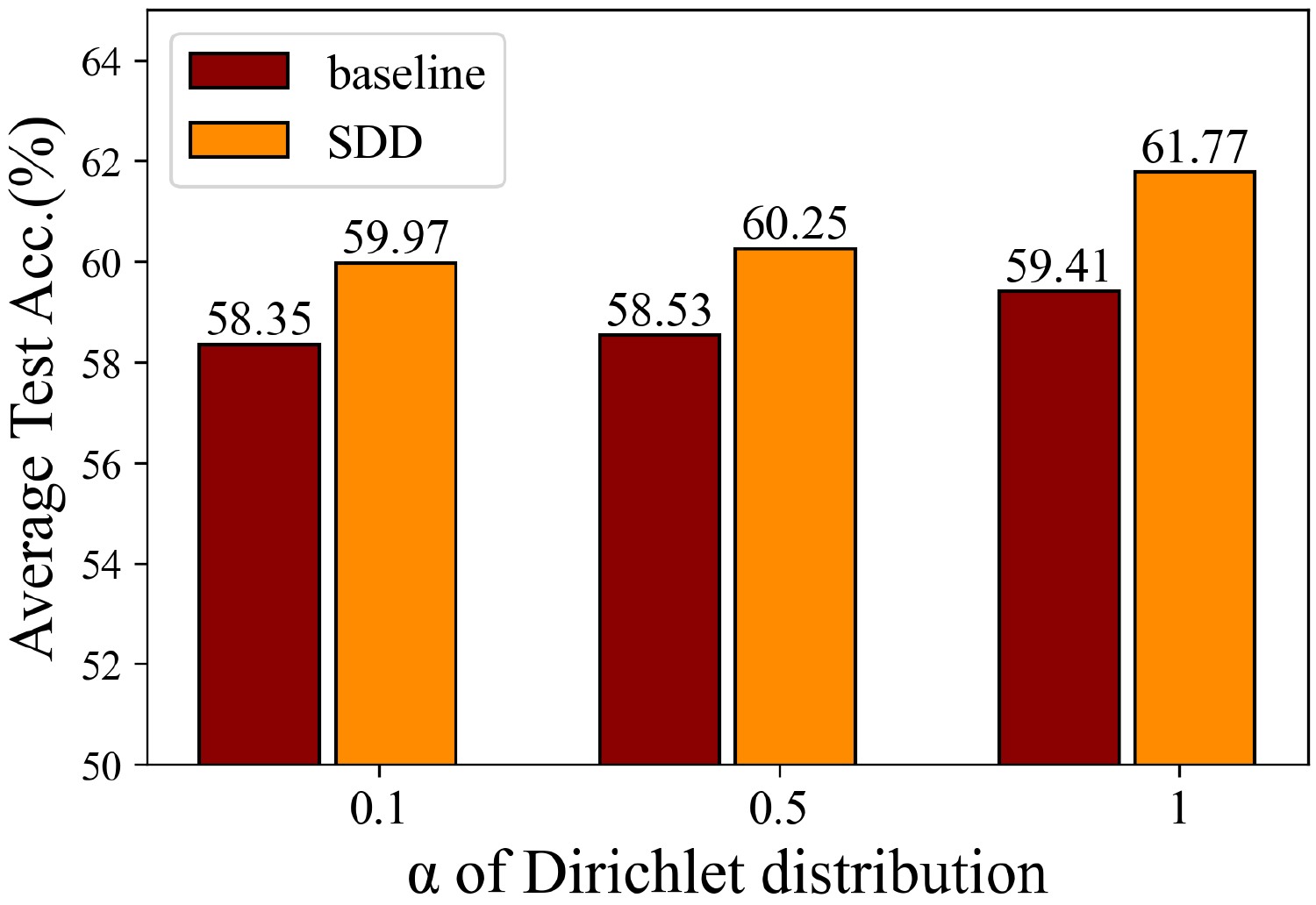}
		\label{fig.7(b)}
	}
	\caption{Performance of the proposed SDD and the baseline using different Dirichlet distribution value $\alpha$ on CIFAR100.}
	\label{fig.cifar_dis}
\end{figure} 

\begin{figure}[t]
	\centering
	\subfloat[Final accuracy]{
		\includegraphics[width=0.47\linewidth]{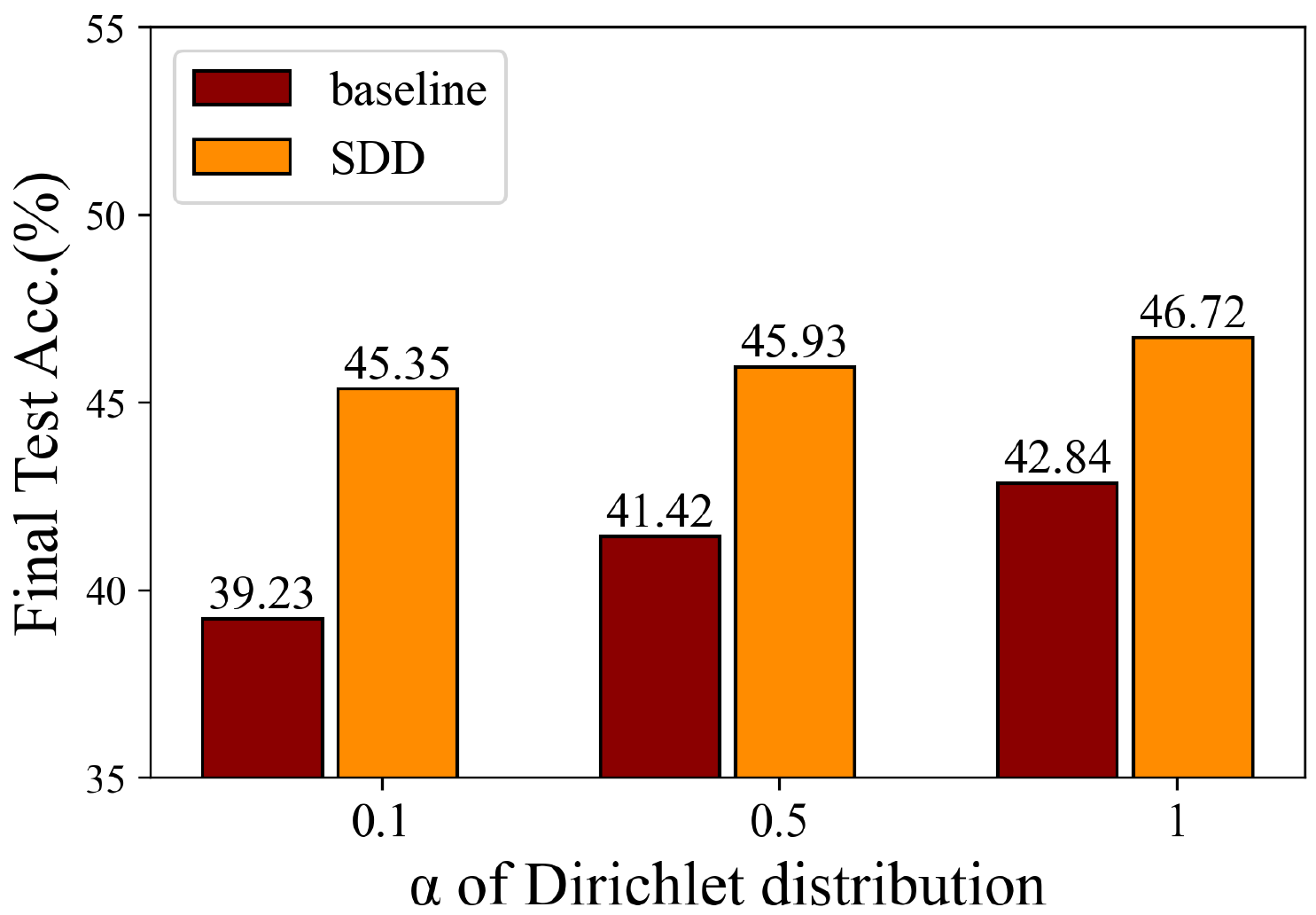}
		\label{fig.8(a)}
	}
	\hfill
	\subfloat[Average accuracy]{
		\includegraphics[width=0.47\linewidth]{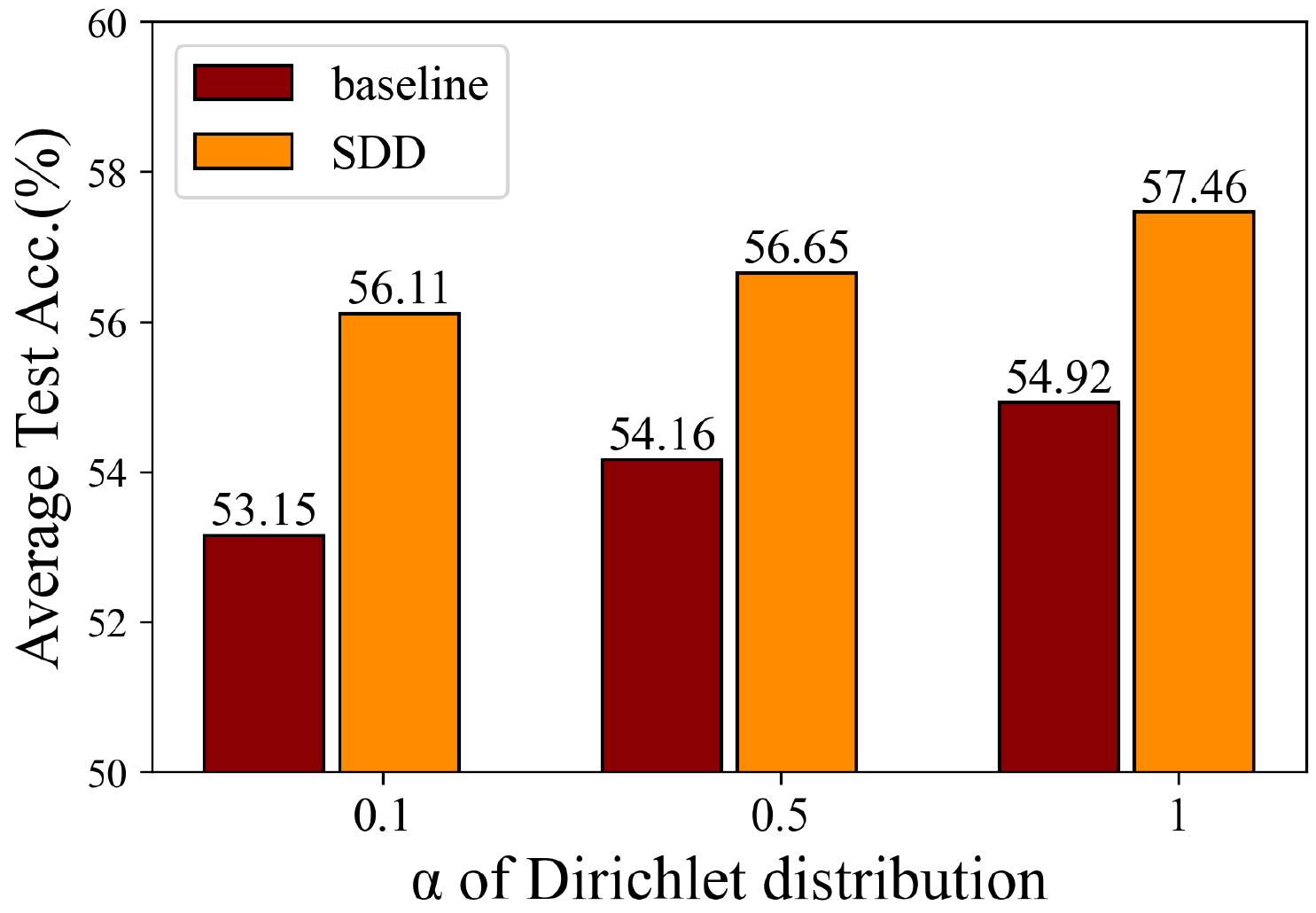}
		\label{fig.8(b)}
	}
	\caption{Performance of the proposed SDD and the baseline using different Dirichlet distribution value $\alpha$ on miniImageNet.}
	\label{fig.mini_dis}
\end{figure}

\subsection{Evaluation under different data distributions}
In federated learning, data heterogeneity across different clients is a common concern.
In our F2SCIL paradigm, data scarcity exacerbates the challenges posed by the non-IID nature of training data.
Following the previous work~\cite{li2022federated}, we distribute the training data to clients by drawing $\alpha \sim \mathrm{Dir} (\alpha \mathbf{p})$ from a Dirichlet distribution, where $\mathbf{p}$ denotes the prior distribution of new classes in a session and $\alpha>0$ controls the non-IID characteristic of training data across multiple clients.
As $\alpha \to 0$, it implies that each client contains training samples from a single class.
Conversely, as $\alpha \rightarrow \infty$, the distribution of training data across clients would perfectly align with the prior class distribution
In the previous experiments, $\alpha$ is set to 1.
Fig.~\ref{distributions} visualizes the distribution of local training data with different Dirichlet distribution value $\alpha$ in an incremental session.

Next, we vary $\alpha$ to adjust the non-IID degree of the training data and evaluate the generalization performance of the proposed SDD across different data distributions.
Fig.~\ref{fig.cifar_dis} and Fig.~\ref{fig.mini_dis} illustrate the experimental results on the CIFAR100 and miniImageNet datasets, respectively.
For the baseline, the data heterogeneity and scarcity result in significant variations among the local models of the clients, leading to a substantial performance decrease of the global model.
Our proposed SDD, using the CSWA strategy, is able to aggregate multiple local models effectively, constructing a more stable global model across different data distributions.

\section{Conclusion}
In this paper, we have undertaken the exploration of the Federated Few-Shot Class-Incremental
Learning (F2SCIL) paradigm.
We introduced a novel Synthetic Data-Driven (SDD) framework, where the Noise-Aware Generative Replay (NAGR) module leverages synthetic replay data to preserve old knowledge during client-side class-incremental learning.
The Class-Specific Weighted Aggregation (CSWA) strategy then consolidates diverse local models into a high-performance global model.
In non-IID settings, the proposed SDD framework has achieved outstanding performance across three datasets.
Ablation studies confirm that the NAGR module effectively addresses noisy pseudo-labels in synthetic replay data, mitigating catastrophic forgetting of previously learned knowledge.
Comparisons with existing aggregation strategies also highlight the superiority of the CSWA approach in managing local model discrepancies due to data heterogeneity and in achieving robust model aggregation.

\bibliographystyle{IEEEtran}
\bibliography{mybib}

\vfill

\end{document}